%% file: main.tex
\documentclass{article}
\usepackage{iclr2027_conference,times}
\usepackage{amsmath,amssymb,graphicx,booktabs,array,tikz,float}
\usepackage[table]{xcolor}
\usepackage{colortbl}
\usepackage{enumitem}
\usepackage{placeins}
\usepackage{algorithm,algorithmic}
\usepackage[hidelinks]{hyperref}
\newcommand{\EIR}{\operatorname{EIR}}
\newcommand{\Pint}{P_{\mathrm{int}}}
\newcommand{\rnom}{\rho_{\mathrm{nom}}}

\usetikzlibrary{arrows,positioning,calc,decorations.pathreplacing}
\iclrfinalcopy
\title{Dude, Where's My State? Execution Information Requirements for Stateful Agents}
\author{Nikita Mehrotra \And Ashish Tiwari \And Priyanshu Gupta \And Sumit Gulwani}
\begin{document}
\maketitle
\input{sections/abstract}
\input{sections/introduction}
\input{sections/execution_information_requirement}
\input{sections/lacuna}

\input{sections/results}
\input{sections/discussion_and_limitations}
\input{sections/related_work}
\input{sections/conclusion}
\bibliography{references}
\bibliographystyle{iclr2027_conference}
\input{sections/submission_statements}
\appendix
\input{appendix/workload_proofs}
\input{appendix/experimental_details}
\input{appendix/results_details}
\input{appendix/deployment_results}

\end{document}

%% file: sections/abstract.tex
\begin{abstract}
Long-running agents must preserve information that later steps depend on.
We introduce the \emph{Execution Information Requirement} (EIR), a lower
bound on the information that must remain accessible for correct
completion under specified task and access conditions. We develop
\emph{LACUNA}, a framework that generates tasks with known dependencies
and varies information demand, retention, and recovery separately from
the difficulty of individual operations. Across four models, restoring
a missing result raises accuracy on affected recall steps to 100\%,
compared with 0\% for equal-length irrelevant information. Sufficient
storage alone does not ensure success: retention policies can discard
required results, errors can propagate through later computations, and
agents can stop before recovery is complete. We also introduce
\emph{VESTIGE}, which uses agent execution traces to construct semantic graphs
and measure information demand for real tasks. Across 72{,}562 software-agent
trajectories, VESTIGE reveals a steeper distance-related decline in
solution-relevant rereading for failed runs (RR 0.951 per distance doubling),
while adjusted peak demand alone is not associated with failure. Together,
these contributions support evaluating whether agents preserve and recover
the information their tasks require.
\end{abstract}

%% file: sections/introduction.tex
\section{Introduction}
\label{sec:introduction}

\begin{figure}[H]
\centering
\begin{minipage}[t]{0.49\linewidth}
\centering
\includegraphics[width=\linewidth]{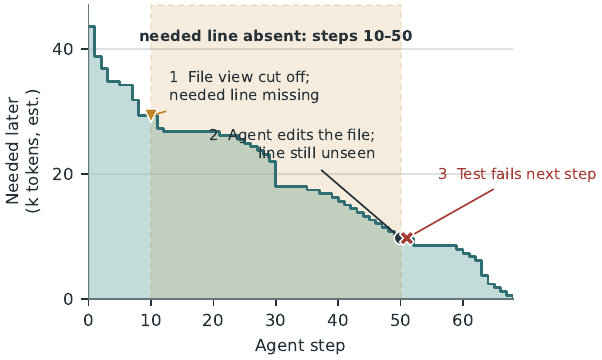}
\par\vspace{-0.8ex}{\scriptsize\textbf{A}}
\end{minipage}\hfill
\begin{minipage}[t]{0.49\linewidth}
\centering
\includegraphics[width=\linewidth]{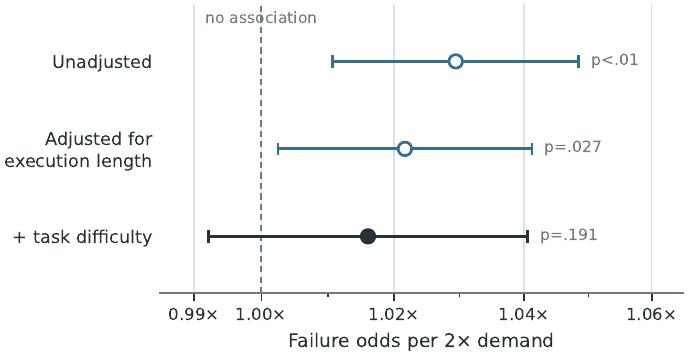}
\par\vspace{-0.8ex}{\scriptsize\textbf{B}}
\end{minipage}
\caption{\textbf{Missing information and task difficulty in real agent tasks.}
\textbf{A:} The agent edits a file without seeing a clipped line; the
next test fails. Teal shows estimated information used later.
\textbf{B:} Higher memory demand is associated with more failures,
but the association is no longer statistically significant after
accounting for task difficulty ($p=.191$). Dots show failure odds
ratios; bars show 95\% confidence intervals. The dashed line at 1
means no association.}
\label{fig:trajectory-clipping-mechanism}
\label{fig:pressure-failure}
\end{figure}

Long-running agents must preserve information across interdependent steps
and recover it when needed. An agent can have ample context capacity and
be capable of its next operation, yet lack the information that operation
requires. Figure~\ref{fig:trajectory-clipping-mechanism}A illustrates the
problem: an OpenHands agent edits a file after a clipped view omits a line
later changed by the reference fix, and the next test fails.

Memory systems use summarization, retrieval, and learned retention
policies to manage information
\citep{packer2023memgpt,kang2025acon,sun2026contextfolding,
zhang2026memact,wu2026proactive}.
Evaluations measure recall, task completion, and memory use
\citep{liu2024lost,vodrahalli2024michelangelo,jimenez2024swe,
yao2024tau,merrill2026terminalbench,shen2026mem2actbench,he2026memoryarena}.
To explain memory-management failures, we also need to establish what
information a task requires and separate the effects of missing information
from the difficulty of the computation. We must also understand the cost
of recovery, since reconstructing one missing result may require repeating
several earlier computations.

We introduce the \emph{Execution Information Requirement} (EIR), a lower
bound in bits on the information from earlier steps that must remain
accessible for correct completion. EIR depends on what later steps may
require and what information they can access, independently of how an
agent represents it. For example, retaining a sum suffices if only that
sum is needed; answering queries about individual values requires more
information. We define \emph{information pressure} as EIR divided by
accessible storage capacity, both measured in the same units. Tasks that
demand more memory also tend to be harder. In 3{,}592 analyzed OpenHands task
executions, peak demand is not significantly associated with failure after
adjustment for execution length and independently estimated task difficulty
(OR 1.016 per demand doubling, $p=.191$; Figure~\ref{fig:pressure-failure}B).

We therefore introduce \emph{LACUNA} to vary how much information a task
requires and what the agent can access, independently of the difficulty
of individual operations. Its six workloads instantiate controlled
aggregation, state-tracking, candidate-maintenance, retrieval, and memory-update
demands. Across four frontier models, restoring
the exact missing result raises accuracy on the affected step to 100\%, compared
with 0\% for matched irrelevant information and 11.3\% on average without
an injection. These interventions establish that unavailable information
causes the tested errors. We evaluate four memory architectures under
increasing information pressure and define \emph{recovery cones} to
describe the earlier results and computations needed to reconstruct a
missing result.

We connect these controlled experiments to real tasks through
\emph{VESTIGE}. It replays an agent's execution over observed environment
state and constructs a semantic action graph connecting actions, accessed or
modified artifacts, and reference-solution activity. This makes implicit
information dependencies measurable when a task provides no dependency graph.
Across 4{,}987 SWE-bench Verified runs, failed executions exhibit a steeper
distance-related decline in rereading solution-relevant files: the interaction
rate ratio is 0.951 per token-distance doubling. Across runs, the same
representation estimates task-demand profiles; within a run, it measures how
efficiently the agent revisits goal-connected state.

Our contributions are:
\begin{itemize}
    \item \textbf{A foundation for studying agent memory.} EIR quantifies
    what information must remain accessible; recovery cones describe
    the computation needed to reconstruct missing results.
    \item \textbf{A controlled evaluation framework.} LACUNA varies
    required and available information independently of the difficulty
    of individual operations, enabling causal tests and comparisons of
    retention and recovery across four models and four memory architectures.
    \item \textbf{VESTIGE.} A semantic action graph that reconstructs
    observable dependencies in real tasks, enabling task-demand profiles and
    per-run analysis of access to goal-connected state.
\end{itemize}

%% file: sections/execution_information_requirement.tex
\section{Execution Information Requirement}
\label{sec:eir}
\label{sec:model}

The information an agent must preserve depends on what the remaining
steps may require. It can discard details that cannot change a required
response, but must preserve the distinctions that can. Following the
communication view of~\citet{kushilevitz1997communication}, we define the
\emph{Execution Information Requirement} (EIR) as the minimum information
that must pass from earlier execution to the remaining work.

\paragraph{An example.}
Figure~\ref{fig:eir-equivalence} shows six exploration histories for a
timezone bug. After the dashed line, the entire remaining task is to
identify the file to edit and whether to return an aware or naive datetime.
Histories $h_1$ and $h_2$ both require $(\texttt{dates.py},\text{aware})$
despite different exploration paths, so they can share a memory
representation. History $h_3$ requires $(\texttt{dates.py},\text{naive})$
and must remain distinguishable. The six histories yield four distinct
answer pairs. Retaining which pair applies is necessary and sufficient,
giving $\operatorname{EIR}_t=\log_2 4=2$ bits.

\input{figures/eir_debugging_rows}

\paragraph{Definition.}
Consider a point after step $t$, which we call an \emph{execution cut}.
It separates the completed history from the remaining work $\mathcal W_t$.
We specify the possible histories, future inputs, required outputs, and
sources the agent can access. We call these assumptions the
\emph{workload contract}.

Fix information $Z_t$ supplied separately, such as the task specification
and public parameters. All other information about the past accessible
after the cut must be represented in the retained state, including
external files and memory that the agent can retrieve.
Let $\mathcal X_t$ contain the histories consistent
with this information at the cut, and let $\mathcal U_t$ be the common set
of allowed future input sequences. For workloads with uniquely specified
outputs, let $F_t(x,u;Z_t)$ be the
required output sequence after history $x$ and future inputs $u$.
Two histories are \emph{continuation-equivalent} if every allowed future
input requires the same output after either history:
\begin{equation}
 x\equiv_{\mathcal W_t,Z_t}x'
 \quad\Longleftrightarrow\quad
 F_t(x,u;Z_t)=F_t(x',u;Z_t)
 \quad\text{for all }u\in\mathcal U_t.
 \label{eq:continuation-equivalence}
\end{equation}
Let $N_t$ be the number of equivalence classes, or groups of histories
with identical future responses. We define
\begin{equation}
 \operatorname{EIR}_t(\mathcal W_t;Z_t)=\log_2 N_t
 \quad\text{bits}.
 \label{eq:eir}
\end{equation}

A sufficient state must distinguish these $N_t$ classes. Otherwise, two
histories requiring different responses would leave the agent in the same
state, making a correct response impossible for at least one of them.
Conversely, the class index determines every required future response.
Thus, for finite $N_t$, the minimum fixed-length binary representation
uses $\lceil\log_2 N_t\rceil$ bits. This argument concerns information;
it does not bound the computation needed to encode or use that state.

\paragraph{What EIR measures.}
EIR is a worst-case, zero-error requirement: correctness must hold for
every allowed history and future input. Any compression is permitted if
it preserves the required distinctions. EIR is measured in bits; its
conversion to tokens depends on the representation.
Appendix~\ref{sec:computing-eir} gives the finite enumeration procedure.

\paragraph{Capacity and availability.}
Let $C_t>0$ be the capacity of the available storage in bits, so it can
represent $2^{C_t}$ distinct states. Using the same access assumptions
as EIR, define \emph{information pressure} as
\begin{equation}
 P_t=\frac{\operatorname{EIR}_t(\mathcal W_t;Z_t)}{C_t}.
 \label{eq:information-pressure}
\end{equation}
If $P_t>1$, the storage cannot distinguish all required classes, so no agent
can guarantee success on every allowed case. If $P_t\leq1$, sufficient
storage is possible, but the agent must still preserve the right
information and perform the computation correctly. For example, if $m$
independent $b$-bit values must be retained and storage holds $h$ such
values, the ratio is $mb/(hb)=m/h$.

EIR specifies what must remain accessible; testing whether missing
information causes errors requires controlling what the agent can access.
Section~\ref{sec:lacuna} introduces LACUNA, our evaluation framework for
this purpose. Section~\ref{sec:results-vestige} introduces VESTIGE for measuring
information demand in real tasks (details in
Appendix~\ref{sec:results-deployment}).

%% file: figures/eir_debugging_rows.tex
\definecolor{eirBlue}{HTML}{2878B5}
\definecolor{eirOrange}{HTML}{D96C24}
\definecolor{eirGreen}{HTML}{3A8F55}
\definecolor{eirPurple}{HTML}{7656A5}
\definecolor{eirInk}{HTML}{26343B}
\definecolor{eirPaper}{HTML}{F5F7F8}
\definecolor{eirGrey}{HTML}{E2E6E8}

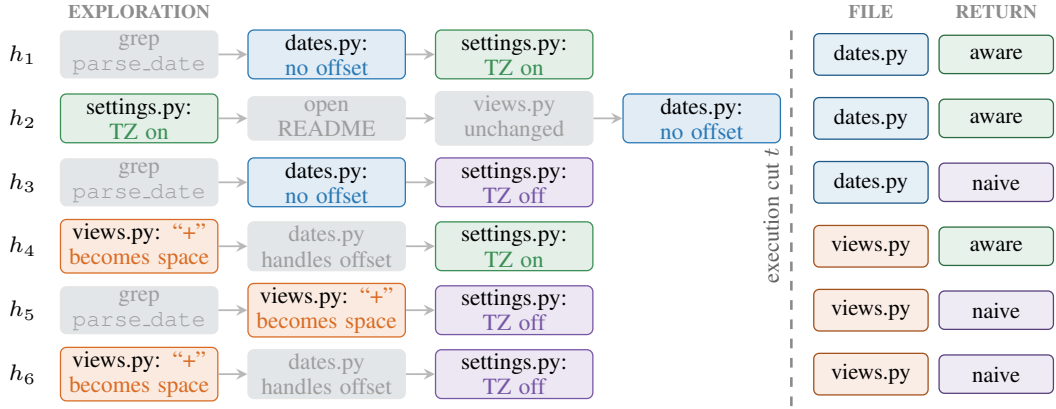
\begin{figure*}[t]
\centering
\resizebox{\textwidth}{!}{%
\begin{tikzpicture}[
  x=1cm,y=1cm,font=\sffamily,
  chip/.style={rounded corners=.65mm,draw=black!20,fill=white,
               minimum width=1.72cm,minimum height=.46cm,text width=1.56cm,
               align=center,font=\fontsize{7.2}{7.8}\selectfont,inner sep=1.2pt},
  irrelevant/.style={chip,draw=black!10,fill=eirGrey,text=black!38},
  answer/.style={rounded corners=.6mm,draw=#1!75!black,fill=#1!12,
                 minimum width=1.25cm,minimum height=.42cm,font=\scriptsize},
  flow/.style={->,>=stealth,line width=.55pt,draw=black!28},
]

\node[font=\tiny\bfseries,text=black!45] at (1.45,4.92) {EXPLORATION};
\node[font=\tiny\bfseries,text=black!45] at (9.45,4.92) {FILE};
\node[font=\tiny\bfseries,text=black!45] at (10.82,4.92) {RETURN};

\node[font=\scriptsize\itshape,anchor=east] at (.45,4.45) {$h_1$};
\node[irrelevant] (h1a) at (1.45,4.45) {grep \texttt{parse\_date}};
\node[chip,draw=eirBlue,fill=eirBlue!14] (h1b) at (3.50,4.45) {dates.py: \textcolor{eirBlue}{no offset}};
\node[chip,draw=eirGreen,fill=eirGreen!14] (h1c) at (5.55,4.45) {settings.py: \textcolor{eirGreen}{TZ on}};
\draw[flow] (h1a)--(h1b); \draw[flow] (h1b)--(h1c);

\node[font=\scriptsize\itshape,anchor=east] at (.45,3.75) {$h_2$};
\node[chip,draw=eirGreen,fill=eirGreen!14] (h2a) at (1.45,3.75) {settings.py: \textcolor{eirGreen}{TZ on}};
\node[irrelevant] (h2b) at (3.50,3.75) {open README};
\node[irrelevant] (h2c) at (5.55,3.75) {views.py unchanged};
\node[chip,draw=eirBlue,fill=eirBlue!14] (h2d) at (7.60,3.75) {dates.py: \textcolor{eirBlue}{no offset}};
\draw[flow] (h2a)--(h2b); \draw[flow] (h2b)--(h2c); \draw[flow] (h2c)--(h2d);

\node[font=\scriptsize\itshape,anchor=east] at (.45,3.05) {$h_3$};
\node[irrelevant] (h3a) at (1.45,3.05) {grep \texttt{parse\_date}};
\node[chip,draw=eirBlue,fill=eirBlue!14] (h3b) at (3.50,3.05) {dates.py: \textcolor{eirBlue}{no offset}};
\node[chip,draw=eirPurple,fill=eirPurple!14] (h3c) at (5.55,3.05) {settings.py: \textcolor{eirPurple}{TZ off}};
\draw[flow] (h3a)--(h3b); \draw[flow] (h3b)--(h3c);

\node[font=\scriptsize\itshape,anchor=east] at (.45,2.35) {$h_4$};
\node[chip,draw=eirOrange,fill=eirOrange!14] (h4a) at (1.45,2.35) {views.py: \textcolor{eirOrange}{``+'' becomes space}};
\node[irrelevant] (h4b) at (3.50,2.35) {dates.py handles offset};
\node[chip,draw=eirGreen,fill=eirGreen!14] (h4c) at (5.55,2.35) {settings.py: \textcolor{eirGreen}{TZ on}};
\draw[flow] (h4a)--(h4b); \draw[flow] (h4b)--(h4c);

\node[font=\scriptsize\itshape,anchor=east] at (.45,1.65) {$h_5$};
\node[irrelevant] (h5a) at (1.45,1.65) {grep \texttt{parse\_date}};
\node[chip,draw=eirOrange,fill=eirOrange!14] (h5b) at (3.50,1.65) {views.py: \textcolor{eirOrange}{``+'' becomes space}};
\node[chip,draw=eirPurple,fill=eirPurple!14] (h5c) at (5.55,1.65) {settings.py: \textcolor{eirPurple}{TZ off}};
\draw[flow] (h5a)--(h5b); \draw[flow] (h5b)--(h5c);

\node[font=\scriptsize\itshape,anchor=east] at (.45,.95) {$h_6$};
\node[chip,draw=eirOrange,fill=eirOrange!14] (h6a) at (1.45,.95) {views.py: \textcolor{eirOrange}{``+'' becomes space}};
\node[irrelevant] (h6b) at (3.50,.95) {dates.py handles offset};
\node[chip,draw=eirPurple,fill=eirPurple!14] (h6c) at (5.55,.95) {settings.py: \textcolor{eirPurple}{TZ off}};
\draw[flow] (h6a)--(h6b); \draw[flow] (h6b)--(h6c);

\draw[dashed,draw=black!46,line width=.8pt] (8.58,.60)--(8.58,4.78);
\node[font=\scriptsize,text=black!62,rotate=90] at (8.36,2.69) {execution cut $t$};

\foreach \y/\file/\filecol/\ret/\retcol in {
  4.45/dates.py/eirBlue/aware/eirGreen,
  3.75/dates.py/eirBlue/aware/eirGreen,
  3.05/dates.py/eirBlue/naive/eirPurple,
  2.35/views.py/eirOrange/aware/eirGreen,
  1.65/views.py/eirOrange/naive/eirPurple,
  .95/views.py/eirOrange/naive/eirPurple}{
  \node[answer=\filecol] at (9.45,\y) {\file};
  \node[answer=\retcol] at (10.82,\y) {\ret};
}

\end{tikzpicture}%
}
\caption{\textbf{Different histories can require the same information.}
The remaining task asks which file to edit and whether to return an
aware or naive datetime. Histories with the same answers form one
group; grey observations do not change either answer.}
\label{fig:eir-equivalence}
\end{figure*}

%% file: sections/lacuna.tex
\section{LACUNA: A Framework for Evaluating Agent Memory}
\label{sec:lacuna}

\textbf{LACUNA} is a framework for generating multi-step tasks and testing
how agents retain and recover information. Tasks are built recursively:
each step combines a new input with results from earlier steps. Specifying
the operations and dependencies lets us generate task instances of varying
length and information demand, with known correct outputs. We then vary
access to earlier results while keeping the task fixed, and test whether
restoring missing information repairs errors.
Figure~\ref{fig:lacuna-framework} shows this evaluation procedure.

\input{figures/lacuna_workflow}

\input{sections/lacuna_design}

%% file: figures/lacuna_workflow.tex
\definecolor{lacunaNavy}{HTML}{28536B}
\definecolor{lacunaTeal}{HTML}{267A68}
\definecolor{lacunaGold}{HTML}{B7791F}
\definecolor{lacunaPlum}{HTML}{76516F}
\definecolor{lacunaRed}{HTML}{B44C4C}
\definecolor{lacunaPaper}{HTML}{F7F9F8}

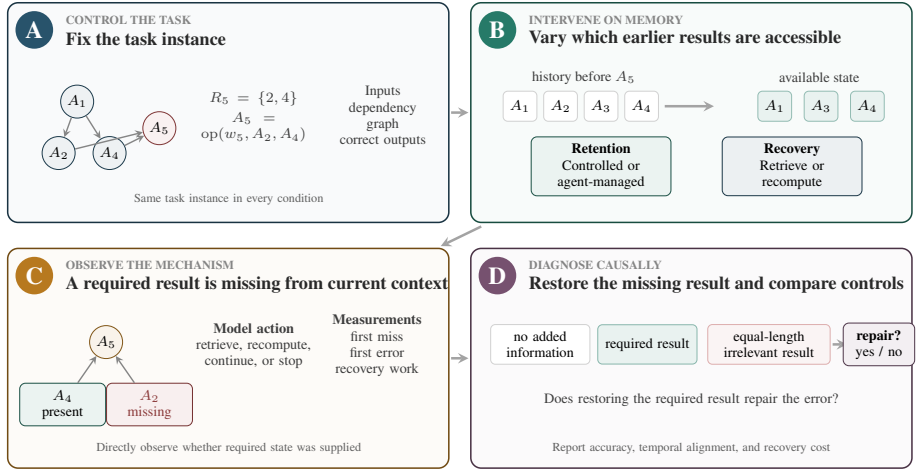
\begin{figure}[t]
\centering
\resizebox{0.86\linewidth}{!}{%
\begin{tikzpicture}[
  x=1cm,y=1cm,font=\sffamily,
  panel/.style={rounded corners=1.8mm,draw=#1!62!black,line width=.75pt,fill=#1!3},
  badge/.style={circle,fill=#1,text=white,font=\bfseries\large,
                minimum size=6.5mm,inner sep=0pt},
  title/.style={font=\bfseries\small,text=black!84,anchor=west},
  kicker/.style={font=\tiny\bfseries,text=black!48,anchor=west},
  body/.style={font=\scriptsize,text=black!76,align=center},
  micro/.style={font=\tiny,text=black!62,align=center},
  chip/.style={rounded corners=.8mm,draw=black!18,fill=white,
               minimum height=5.2mm,font=\scriptsize,align=center},
  state/.style={rounded corners=1mm,draw=#1!65!black,fill=#1!9,
                minimum height=7mm,font=\scriptsize,align=center},
  result/.style={circle,draw=lacunaNavy!70!black,fill=lacunaNavy!8,
                 minimum size=5.7mm,inner sep=0pt,font=\scriptsize},
  flow/.style={->,>=stealth,line width=1.25pt,draw=black!36},
  dep/.style={->,>=stealth,line width=.7pt,draw=black!48}
]

\path[panel=lacunaNavy] (0,4.25) rectangle (7.62,8.05);
\path[panel=lacunaTeal] (8.00,4.25) rectangle (15.62,8.05);
\path[panel=lacunaGold] (0,0) rectangle (7.62,3.80);
\path[panel=lacunaPlum] (8.00,0) rectangle (15.62,3.80);

\draw[flow] (7.66,6.15)--(7.96,6.15);
\draw[flow] (8.18,4.18)--(7.48,3.84);
\draw[flow] (7.66,1.90)--(7.96,1.90);

\node[badge=lacunaNavy] at (.46,7.58) {A};
\node[kicker] at (.88,7.76) {CONTROL THE TASK};
\node[title] at (.88,7.43) {Fix the task instance};

\node[result] (a1) at (1.20,6.34) {$A_1$};
\node[result] (a2) at (.88,5.45) {$A_2$};
\node[result] (a4) at (1.76,5.45) {$A_4$};
\node[result,draw=lacunaRed!70!black,fill=lacunaRed!8] (a5) at (2.62,5.88) {$A_5$};
\draw[dep] (a1)--(a2); \draw[dep] (a1)--(a4);
\draw[dep] (a2)--(a5); \draw[dep] (a4)--(a5);

\node[body,text width=2.15cm] at (4.25,6.08)
  {$R_5=\{2,4\}$\\[2pt]$A_5=\mathrm{op}(w_5,A_2,A_4)$};
\node[body,text width=1.72cm] at (6.48,6.08)
  {Inputs\\dependency graph\\correct outputs};
\node[micro,text width=6.2cm] at (3.82,4.66)
  {Same task instance in every condition};

\node[badge=lacunaTeal] at (8.46,7.58) {B};
\node[kicker] at (8.88,7.76) {INTERVENE ON MEMORY};
\node[title] at (8.88,7.43) {Vary which earlier results are accessible};

\node[body] at (9.95,6.76) {history before $A_5$};
\foreach \x/\lab in {8.85/1,9.55/2,10.25/3,10.95/4}{
  \node[chip,minimum width=.48cm] at (\x,6.26) {$A_{\lab}$};
}
\draw[flow] (11.35,6.26)--(12.20,6.26);
\node[body] at (14.02,6.76) {available state};
\foreach \x/\lab in {13.25/1,14.05/3,14.85/4}{
  \node[chip,minimum width=.57cm,fill=lacunaTeal!10,draw=lacunaTeal!50] at (\x,6.26) {$A_{\lab}$};
}
\node[state=lacunaTeal,text width=2.18cm] at (10.25,5.22)
  {\textbf{Retention}\\Controlled or agent-managed};
\node[state=lacunaNavy,text width=2.18cm] at (13.55,5.22)
  {\textbf{Recovery}\\Retrieve or recompute};

\node[badge=lacunaGold] at (.46,3.33) {C};
\node[kicker] at (.88,3.51) {OBSERVE THE MECHANISM};
\node[title] at (.88,3.18) {A required result is missing from current context};

\node[result,draw=lacunaGold!70!black,fill=lacunaGold!9] (c5) at (1.70,2.18) {$A_5$};
\node[state=lacunaTeal,text width=1.25cm] (c4) at (.95,1.10)
  {$A_4$\\present};
\node[state=lacunaRed,text=lacunaRed!78!black,text width=1.25cm] (c2) at (2.45,1.10)
  {$A_2$\\missing};
\draw[dep] (c4)--(c5); \draw[dep] (c2)--(c5);

\node[body,text width=2.30cm] at (4.25,2.12)
  {\textbf{Model action}\\retrieve, recompute, continue, or stop};
\node[body,text width=1.95cm] at (6.38,2.12)
  {\textbf{Measurements}\\first miss\\first error\\recovery work};
\node[micro,text width=6.4cm] at (3.82,.34)
  {Directly observe whether required state was supplied};

\node[badge=lacunaPlum] at (8.46,3.33) {D};
\node[kicker] at (8.88,3.51) {DIAGNOSE CAUSALLY};
\node[title] at (8.88,3.18) {Restore the missing result and compare controls};

\node[chip,text width=1.48cm,minimum height=.68cm] at (9.20,2.14)
  {no added information};
\node[chip,text width=1.48cm,minimum height=.68cm,
      fill=lacunaTeal!10,draw=lacunaTeal!55] at (11.05,2.14)
  {required result};
\node[chip,text width=1.88cm,minimum height=.68cm,
      fill=lacunaRed!6,draw=lacunaRed!45] at (13.15,2.14)
  {equal-length irrelevant result};
\draw[flow] (14.25,2.14)--(14.50,2.14);
\node[state=lacunaPlum,text width=1.0cm] at (15.05,2.14)
  {\textbf{repair?}\\yes / no};

\node[body,text width=6.2cm] at (11.81,1.18)
  {Does restoring the required result repair the error?};
\node[micro,text width=6.4cm] at (11.81,.34)
  {Report accuracy, temporal alignment, and recovery cost};

\end{tikzpicture}%
}
\caption{\textbf{LACUNA tests whether missing information causes errors.}
Fix a task instance (A), vary access to earlier results (B), and
observe the agent when a required result is missing from context (C).
Compare restoring that result with no added information and an
equal-length irrelevant result (D).}
\label{fig:lacuna-framework}
\end{figure}

%% file: sections/lacuna_design.tex
\subsection{Task construction}
\label{sec:lacuna-design}

Each task is a directed acyclic graph: nodes are steps, and an edge
$r\rightarrow t$ means that step $t$ uses the result of step $r$. Given
new input $w_t$ and identifiers $R_t$ naming earlier results, the agent
computes
\begin{equation}
 A_t=\operatorname{op}\!\left(w_t,(A_r)_{r\in R_t}\right),
 \qquad R_t\subseteq\{1,\ldots,t-1\}.
 \label{eq:lacuna-node}
\end{equation}
Identifiers distinguish results even when their values coincide.

A \emph{workload} specifies how inputs and dependencies are generated,
how results are computed, and which outputs are scored. It also defines
when dependencies are revealed and what may be retained. In online
workloads, the agent learns a step's dependencies only when that step
arrives. A \emph{task instance} is generated with a fixed random seed,
fixing its inputs, dependencies, and correct outputs before execution.
We compare memory conditions on the same instances.

Generation is controlled by the number of steps $n$, operation
$\mathrm{op}$, dependency-window size $m$, maximum lookback $\ell$
(default $m$), and maximum dependencies per step $f_{\mathrm{in}}$.
Each result record contains $s$ value slots with $d$ decimal digits per
slot. These parameters control task length, dependency structure, and
the amount of information in each result.

\paragraph{Workloads used in this study.}
We evaluate six examples of this construction
(Table~\ref{tab:lacuna-workloads}), spanning \emph{dependency aggregation}
(Stepwise Sum), \emph{sufficient-state tracking} (Running Maximum),
\emph{interdependent state tracking} (Stepwise Maximum), \emph{candidate-set
maintenance} (Stepwise Top-$k$), \emph{delayed key--value retrieval} (Full
Lookup), and \emph{interleaved memory update and recall} (Store--Recall).
These are descriptive, not standardized labels; parenthesized operator names
are the formal identifiers. Running Maximum needs one summary value, whereas
Stepwise Maximum preserves reusable intermediate maxima, illustrating how future
use determines what must be retained. Other operations and dependencies fit the
same construction.

We vary the record limit for five workloads and use Running Maximum as
a one-value control. Exact EIR for Stepwise Maximum depends on the
ancestor relationships among results still needed by future steps
(Appendix~\ref{sec:app-eir-calculations}). Stepwise Top-$k$ EIR is not
characterized, so we report its record limits without an exact
EIR-to-capacity ratio.

\begin{table}[t]
\centering
\scriptsize
\caption{\textbf{Controlled information-demand patterns.} Descriptive labels
precede formal workload identifiers. Rules and EIR are for the evaluated
settings. Here $b=\log_2 q$
for a value domain of size $q$, and $\dagger$ marks EIR that depends on
the dependency graph.}
\label{tab:lacuna-workloads}
\input{tables/tab_operators}
\end{table}

Table~\ref{tab:lacuna-workloads} reports the evaluated settings, not universal
workload formulas. At saturated $m=16$ cuts, Stepwise Sum requires $16b$
(generally $|W_t|b$), while Store--Recall requires $8b$ after STORE steps and
$7b$ after RECALL steps. Query and recall identifiers are revealed only when
their step arrives. Stepwise Top-$k$ record-limit results are descriptive
because its numeric EIR is not characterized; Appendix~\ref{sec:workload-proofs}
gives the formal scope.

\subsection{Controlling information availability and recovery}
\label{sec:recovery-control}

Once the task is fixed, we vary what the agent can retain and recover.
We distinguish results needed by future steps (\emph{required state}),
results preserved by the memory system (\emph{retained state}), and
information accessible through the permitted interfaces and recovery
budgets (\emph{available state}). A stored result may require retrieval,
while a result in context may no longer be needed.

The limit $h$ specifies how many complete labelled result records may be
available at once. Under \emph{Controlled Retention}, the harness resets
model-visible context at each step and supplies the most recent $h$
submitted results, or all results if fewer exist. Under
\emph{Agent-Managed Retention}, the agent system chooses what to retain,
summarize, store externally, and retrieve through its permitted interfaces.

Conditions also specify whether missing results can be retrieved directly
or reconstructed. The budget $B_{\mathrm{oracle}}$ limits the number of
directly retrieved results; $B$ limits the number of recomputed steps.
Reconstructing a result may require recovering its dependencies first;
these form its \emph{recovery cone}. A result is unavailable if it cannot
be obtained through the permitted interfaces within these budgets.

\subsection{Evaluation controls}

The controls first establish that the agent can perform the computation,
then test whether missing information causes errors.

\paragraph{Computational capability.}
The \emph{full-information control} supplies all required earlier results
and disables recovery. For window-based workloads under Controlled
Retention, $h=m=\ell$ covers the eligible predecessor window. Remaining
errors reflect computation, use of the wrong result, formatting, or
protocol handling. We attribute errors to information loss only in
conditions that pass this control.

\paragraph{Causal intervention.}
For a step error, we compare no added information, restoration of the
required result, and an irrelevant value of the same length. If only
restoration repairs the answer, the comparison identifies the missing
information as the cause rather than the added token count. All controlled
prompts fit within the model's context window; reducing $h$ shortens the
input while increasing the demand relative to available storage.

\paragraph{Errors and measurements.}
We distinguish an incorrect answer with correct inputs available
(\emph{computation error}), an unavailable required result
(\emph{state-supply error}), and reuse of an incorrect earlier result
(\emph{cascade error}). An \emph{execution failure} is an incorrect final
task outcome. LACUNA records required, retained, and available results;
retrievals and recomputed steps; recovery depth; latency; and subsequent
errors. We report both step and final-task accuracy to capture how
intermediate mistakes affect task completion.

\paragraph{Interface checks.}
We check that the retention interface does not carry information between
steps through unintended channels. Across 18,068 checks on 100 generated
instances, the audit finds no violations in payload width, identifiers,
occupancy, ordering, serialization, message length, or isolation between
context resets. The appendix provides the full audit, experiment
configurations, model settings, and trial counts.

%% file: tables/tab_operators.tex
\par
\renewcommand{\arraystretch}{1.05}
\setlength{\tabcolsep}{1.5pt}
\tikzset{
	lacuna prior/.style={circle,draw=black!45,fill=black!18,minimum size=5.5pt,inner sep=0pt},
	lacuna target/.style={circle,fill=black!65,minimum size=6.5pt,inner sep=0pt},
	lacuna edge/.style={->,>=stealth,thick,draw=black!70},
	lacuna action/.style={circle,draw=black!65,fill=white,minimum size=11pt,inner sep=0pt,font=\tiny\bfseries}
}
\begin{tabular}{@{}>{\raggedright\arraybackslash}p{0.19\linewidth}>{\raggedright\arraybackslash}p{0.22\linewidth}>{\centering\arraybackslash}p{0.17\linewidth}>{\raggedright\arraybackslash}p{0.21\linewidth}>{\centering\arraybackslash}p{0.12\linewidth}@{}}
\textbf{Demand pattern (workload)} & \textbf{Operation} & \textbf{Dependencies} & \textbf{Evaluated outputs} & \textbf{EIR} \\
\midrule
\cellcolor{blue!5}\textbf{Dependency aggregation} (\emph{Stepwise Sum}) & Modular sum over two dependencies sampled from a preceding window &
\tikz[baseline=-0.5ex,x=0.50cm,y=0.34cm]{\node[lacuna prior] (a) at (0,0.5) {}; \node[lacuna prior] (b) at (0,-0.5) {}; \node[lacuna target] (c) at (1,0) {}; \draw[lacuna edge] (a)--(c); \draw[lacuna edge] (b)--(c);} & Every step; identifiers revealed only when used & $16b$ \\
\cellcolor{blue!5}\textbf{Sufficient-state tracking} (\emph{Running Maximum}) & Update the maximum along a fan-in-one chain &
\tikz[baseline=-0.5ex,x=0.45cm]{\node[lacuna prior] (a) at (0,0) {}; \node[lacuna prior] (b) at (1,0) {}; \node[lacuna prior] (c) at (2,0) {}; \node[lacuna target] (d) at (3,0) {}; \draw[lacuna edge] (a)--(b); \draw[lacuna edge] (b)--(c); \draw[lacuna edge] (c)--(d);} & Final global maximum & $b$ \\
\cellcolor{blue!5}\textbf{Interdependent state tracking} (\emph{Stepwise Maximum}) & Take the maximum over two randomly selected earlier results &
\tikz[baseline=-0.5ex,x=0.50cm,y=0.34cm]{\node[lacuna prior] (a) at (0,0.5) {}; \node[lacuna prior] (b) at (0,-0.5) {}; \node[lacuna target] (c) at (1,0) {}; \draw[lacuna edge,bend left=12] (a) to (c); \draw[lacuna edge,bend right=12] (b) to (c);} & Every intermediate local maximum & graph-dependent$^{\dagger}$ \\
\cellcolor{blue!5}\textbf{Candidate-set maintenance} (\emph{Stepwise Top-$k$}) & Compute the top-$k$ values over two earlier results &
\tikz[baseline=-0.5ex,x=0.50cm,y=0.34cm]{\node[lacuna prior] (a) at (0,0.5) {}; \node[lacuna prior] (b) at (0,-0.5) {}; \node[lacuna target] (c) at (1,0) {}; \draw[lacuna edge] (a)--(c); \draw[lacuna edge] (b)--(c);} & Every frontier after warm-up & not characterized \\
\cellcolor{blue!5}\textbf{Delayed key--value retrieval} (\emph{Full Lookup}) & Store labelled values, then query arbitrary labels; 16 STORE + 48 QUERY steps &
\tikz[baseline=-0.5ex,x=0.70cm]{\node[lacuna action] (s) at (0,0) {S}; \node[lacuna action] (q) at (1.3,0) {Q}; \draw[lacuna edge] (s)--(q);} & Every query & $mb$ \\
\cellcolor{blue!5}\textbf{Interleaved memory update and recall} (\emph{Store--Recall}) & Alternate STORE and RECALL steps over eligible stored identifiers &
\tikz[baseline=-0.5ex,x=0.70cm]{\node[lacuna action] (s) at (0,0) {S}; \node[lacuna action] (r) at (1.3,0) {R}; \draw[lacuna edge] (s)--(r);} & Every recall & $7b$--$8b$ \\
\bottomrule
\end{tabular}

%% file: sections/results.tex
\section{Results}
\label{sec:results}

We test the effects of missing information, recovery cost, retention
behavior, and memory architectures. Information-loss claims use conditions
that pass the full-information control; residual computation errors in
Stepwise Top-$k$ are analyzed separately. Detailed comparisons and additional
figures appear in Appendix~\ref{sec:app-results-details}.

\subsection{RQ1: Does unavailable required state predict and cause execution errors?}
\label{sec:results-pressure}

\begin{figure}[t]
\centering
\includegraphics[width=0.88\linewidth]{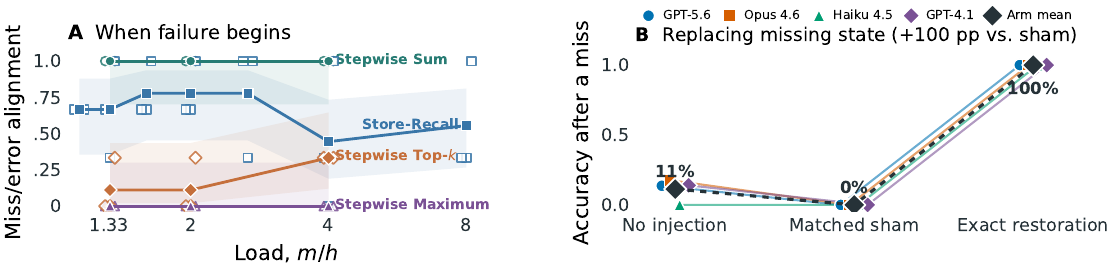}
\caption{\textbf{Missing inputs predict errors; restoring them repairs
answers.} \textbf{A:} Fraction of trials whose first missing dependency
coincides with the first error. Bands show 95\% confidence intervals.
\textbf{B:} Restoring the required result repairs affected Store--Recall
answers; adding an irrelevant value of the same length does not.}
\label{fig:temporal-alignment}
\label{fig:rq1-causal-rescue}
\end{figure}

\textbf{Restoring missing information repairs errors.} In Store--Recall
($m=8$, $h=4$, recovery disabled), supplying the required result raises
accuracy on affected recall steps to 100\% across four models, compared
with 0\% for an equal-length irrelevant value and 11.3\% without added
information (Figure~\ref{fig:rq1-causal-rescue}B). The intervention
isolates information content from added prompt length. A Stepwise Maximum
replication also reaches 1.000 with exact restoration, versus 0.642
without restoration and 0.714 with irrelevant information
(Appendix Table~\ref{tab:rq1-stepwise-max-rescue}).

The first missing dependency coincides with the first error in all 27
Stepwise Sum trials with restricted memory. Alignment is weaker when a
missing value can be guessed or does not change the maximum. Across
96 Full Lookup trials, query accuracy falls from 0.997 at $h=16$ to
0.092 at $h=2$, while queries for retained values remain nearly perfect.
Errors can also propagate: one 64-step Stepwise Sum execution contains
19 errors at missing-input steps and 26 cascade errors, with no independent
computation errors. Later steps correctly apply the operation to already
incorrect inputs. Appendix~\ref{sec:results-rq1} reports the retention
curves and error decomposition.

\subsection{RQ2: What determines reconstruction cost?}
\label{sec:results-recovery}

\textbf{Recovery work follows the missing dependencies.} Direct retrieval
costs one read. Reconstruction with reuse evaluates each node in the
recovery cone once; without reuse, shared ancestors are evaluated repeatedly.
Mean work on the tested chain, balanced, and bushy graphs is 13.0, 38.5,
and 45.3 node evaluations with reuse. Without reuse, the bushy-graph mean
rises to approximately $1.1\times10^{10}$. Cone size measures total work;
depth describes the longest sequential dependency chain.

In 88 agent-executed probes, the model selects reads while the harness
performs arithmetic. All 69 successful recoveries use exactly the minimum
number of reads; the other 19 stop before reconstructing the target.
Thus, the observed failures involve incomplete recovery rather than
redundant traversal. Appendix~\ref{sec:results-rq2} provides the graph
comparisons and per-workload results.

\subsection{RQ3: How does retention behavior change under information pressure?}
\label{sec:results-adaptation}

\textbf{Adaptive retention benefits from repeated queries, while storage
format affects reliability.} In Full Lookup with Zipf-distributed queries
and $h=4$, hit rates are 0.493 for least-frequently-used retention (LFU),
0.438 for least-recently-used retention (LRU), and 0.111 for the fixed
policy. Uniform queries show no adaptive advantage. Direct retrieval
keeps query accuracy at least 0.998, so hit rate exposes the policy difference.

On the agent-managed Store--Recall workload, model-directed scratch storage improves over LRU for Claude Opus 4.6
(0.762 versus 0.664) and GPT-5.6 (0.750 versus 0.650), but reduces accuracy
for Claude Haiku 4.5 (0.438 versus 0.616). Holding LRU fixed, key--value
storage improves accuracy over text scratch storage by 25.9--33.0
percentage points across all four models. Storage format provides the
more consistent benefit in these tests (Appendix~\ref{sec:results-rq3}).

All 96 Stepwise Maximum outcomes are complete: pooled accuracy is 0.654 for
the no-recovery floor, 0.223 for Scratch, 0.222 for LRU, and 0.072 for
Scratch--LRU, so predominantly just-in-time retrieval does not improve this
random-DAG workload (Appendix Table~\ref{tab:rq3-stepwise-max-behaviors}).

\subsection{RQ4: How closely do architectures approach the EIR floor?}
\label{sec:results-policies}

\textbf{Enough capacity does not ensure correct retention.} At $h=16$,
Stepwise Sum can retain all 16 eligible results, yet LRU scores 0.410
compared with 0.934 for exact-window retention. Restoring LRU's missing
results raises its score to 0.934; irrelevant information reaches only
0.444. The restoration test identifies discarded information as the source
of this gap (Appendix Table~\ref{tab:rq2-agg-capacity}). Offline comparisons
use future-aware Belady as a hindsight benchmark, distinct from the
worst-case online requirement defined by EIR.

Complete architectures also depend on correct retrieval and execution.
On Store--Recall, unbounded forced paging reaches 1.000, Graphiti 1.000,
Mem0 0.996, A-MEM 0.959, FIFO 0.811, and Letta 0.370. These are system-level results under different
storage conditions, not equal-capacity policy comparisons. Stepwise
Top-$k$ remains difficult even with forced paging (0.501), consistent
with its computation limitation. Appendix~\ref{sec:results-rq4} separates
policy comparisons from end-to-end architecture results.

\subsection{VESTIGE: What can be measured in real tasks?}
\label{sec:results-vestige}

Real agent logs expose actions and environment changes but not an explicit
dependency graph. VESTIGE replays each execution over the observed environment
and constructs a semantic action graph connecting actions to artifacts they
access or modify and, where available, to reference-solution activity. On 36
synthetic cases with known dependencies and an injected distractor, it retains
all required files and excludes the distractor in every case.

We apply VESTIGE to 72{,}562 software-agent trajectories. Mean estimated peak
demand is 9.0k tokens on SWE-bench Verified, 73.1k on CoderForge, and 4.5k on
OpenHands; reference-file access coverage is 92.0\%, 99.0\%, and 57.4\%.
In the 4{,}987-run reread analysis, each doubling of token distance multiplies
the failed-versus-resolved reread-rate ratio for solution-relevant files,
relative to other files, by 0.951 (95\% CI $[0.911,0.992]$, $p=.020$). Failed
runs therefore show a steeper distance-related decline in solution-relevant
rereading. Peak demand itself is not significantly associated with failure in
the 3{,}592-run OpenHands cohort after adjustment for execution length and
independently estimated task difficulty (OR 1.016 per demand doubling, 95\% CI
$[0.992,1.041]$, $p=.191$). Appendix~\ref{sec:results-deployment} reports
coverage, sensitivities, and cross-domain adapters in their native units.

%% file: sections/discussion_and_limitations.tex
\section{Discussion and Limitations}
\label{sec:discussion}

\paragraph{What the framework makes measurable.}
EIR describes what the remaining computation requires, independently of how a
memory system stores, retrieves, or recomputes it. LACUNA then separates
retention-policy errors from end-to-end failures involving model outputs, tool
actions, or computation. EIR is a worst-case lower bound, not a per-run failure
predictor; outcomes also depend on retention, retrieval, and model capability.

\paragraph{Implications for memory systems.}
The Mem0, Letta, A-MEM, and Graphiti results illustrate why memory quality cannot be inferred from
the presence of an external store alone. Mem0 nearly matches the forced-paging
upper bound on Store--Recall; Graphiti and A-MEM also score 1.000 and 0.959,
whereas Letta performs substantially worse in the same
agent-executed setting. These runs do not establish a universal ranking of
memory products: they combine each system's interface, the model's decisions,
and answer submission. They do show that task-aligned retention and reliable
access are measurable properties, and that a system can fail even when storage
is available.

\paragraph{What VESTIGE enables.}
VESTIGE reconstructs semantic action graphs when real tasks lack declared
dependencies, measuring goal-connected artifacts, liveness, and reuse. This
supports cross-run demand profiles and per-run access-efficiency analyses across
72{,}562 SWE, 318 $\tau$-Bench, and 166 Terminal-Bench trajectories. Its scores
describe observed goal-connected use, not the minimum required by every correct
solution. The traces also reveal clipped tool outputs, but exported logs cannot
always identify whether omitted lines were required. VESTIGE therefore enables
targeted retention and harness experiments based on task demand rather than
context occupancy alone.

\paragraph{Design implications and limitations.}
Memory systems should retain state by predicted future use, track what leaves
resident memory, and price retrieval by dependency work. LACUNA isolates
information pressure but omits some realistic content priors and per-step
difficulty. Exact EIR requires a workload contract and assumes unrestricted
encoding, whereas experiments use fixed-width records. Capacity calibration is
therefore workload-specific; native systems without a common auditable unit are
reported separately with effective counts. Real traces also mix memory,
capability, exploration, and tool-use failures. Future work should address
stochastic workloads, richer tools, and learned recovery policies.

%% file: sections/related_work.tex
\section{Related Work}

\paragraph{Predictive and task-relevant state.}
Several theories define a compact state by asking which pasts have the same
future consequences. Myhill-Nerode equivalence merges prefixes that admit the
same continuations~\citep{nerode1958linear}. Computational mechanics similarly
groups histories that give the same distribution over future observations into
\emph{causal states}~\citep{crutchfield1989inferring,
shalizi2001computational}. EIR uses the same basic idea: merge pasts that the
future cannot distinguish. It applies this idea at a finite execution boundary
with a specified remaining workload. It measures the number of states needed to
guarantee correct continuation in the worst case. In contrast, the information
bottleneck and predictive-information frameworks optimize average-case mutual
information under a probability distribution~\citep{tishby1999information,
bialek2001predictive}. EIR therefore requires no distribution over execution
prefixes: the future computation itself determines which prefixes must remain
distinct.

\paragraph{Long-context evaluation and agent memory.}
Long-context benchmarks reveal that models can fail even when the required
information is present: accuracy varies with evidence position and with how
reasoning is distributed across the context~\citep{liu2024lost,
vodrahalli2024michelangelo}. However, their aggregate recall or task-accuracy
scores do not identify why a failure occurred. They cannot generally distinguish
insufficient memory capacity from retaining the wrong information, failing to
retrieve retained information, or failing to use information that was available.
Prior work tests aggregation and key--value retrieval~\citep{hsieh2024ruler},
knowledge updates~\citep{wu2025longmemeval}, active memory utilization
\citep{shen2026mem2actbench}, and interdependent state tracking
\citep{he2026memoryarena}. LACUNA adopts these as descriptive labels, not a
standardized taxonomy, for controlled workloads with known requirements.
Evaluations of external storage, hierarchical memory, trajectory summaries, and
learned retention policies face the same attribution problem
\citep{park2023generative,packer2023memgpt,kang2025acon,
sun2026contextfolding,zhang2026memact}: an improved final score shows that a
memory mechanism helps, but not which state-availability error it prevents. More targeted
evaluations test long-run state consistency, detect disagreement between an
agent and its environment, separate stages of memory use, or test whether
remembered information supports later actions
\citep{samiei2026illusion,song2026worldmodel,zhu2026aging,
shen2026mem2actbench,he2026memoryarena}. These evaluations localize memory
problems, but do not jointly identify the information required at a failed step
and test whether its absence caused the failure. LACUNA makes
these causes separately measurable by controlling what information later steps
need, what information the agent can access, and whether missing information can
be reconstructed. LACUNA therefore lets developers stress-test memory systems
and policies under controlled memory requirements. Our retrospective demand proxy
uses estimates of these requirements to assess state availability
in completed real-agent runs.

\paragraph{Context compaction and task-aware memory.}
Recent theory studies what a bounded memory should keep. Context-compaction
work relates the information retained in a summary to communication complexity
\citep{tirmazi2026compaction}, while hierarchical and rate-distortion
formulations study how memory can trade representation size against downstream
performance~\citep{talebirad2026hierarchical,colaco2026compaction}. Related
decision-centric work keeps distinctions between histories only when merging
them would reduce future decision quality~\citep{zou2026decision}. Together,
these approaches motivate task-aware memory: a useful representation
should keep the parts of the past that matter to future behavior. They focus
on mechanisms for constructing such representations or on the lossy trade-offs
between memory size and downstream performance. Their evaluations can show that
a memory method improves performance, but do not directly measure whether a
particular failure arose from unavailable required information. EIR specifies
what information the task requires, and LACUNA uses it to create and diagnose
state-unavailability errors while trajectory length, graph structure,
and per-step difficulty are held fixed.

\paragraph{Recomputation and information lower bounds.}
Compiler liveness marks values that may be reused~\citep{kildall1973unified},
while pebbling and rematerialization characterize storage--recomputation
trade-offs~\citep{paul1977space,hong1981io,chen2016sublinear,
gruslys2016memory}. These ideas motivate our recovery-cost analysis. Streaming
and communication lower bounds support our requirement results
\citep{alon1999space,miltersen1998data,kushilevitz1997communication}: arbitrary
labelled access grows with input size, while aggregate and extremum workloads
can use constant-size state.

%% file: sections/conclusion.tex
\section{Conclusion}

Together, EIR, LACUNA, and VESTIGE connect task demand to retention, recovery,
and failures in real agent executions.

%% file: sections/submission_statements.tex
\section*{AI Use Statement}
Generative AI tools were used for manuscript editing, writing, finalization,
and software implementation.

\section*{Reproducibility Statement}
The appendix states workload contracts, proofs, controls, experiment settings,
and observational limitations. The anonymous supplementary archive contains the
LACUNA and trajectory-analysis code, tests, compact aggregate results, theory
artifacts, figure scripts, checksums, and instructions for validation. Raw public
trajectories are not redistributed; the archive identifies their public dataset
sources and the scripts used to process them.

%% file: appendix/workload_proofs.tex
\section{Proofs for Exact EIR Workloads}
\label{sec:workload-proofs}

Let $\mathcal A=\{0,\ldots,q-1\}$ be the value alphabet and let
$b=\log_2 q$.  The arguments below count continuation-equivalence classes
under the workload contracts used in the paper and establish the exact EIR
results reported in Table~\ref{tab:lacuna-workloads}.

\input{appendix/computing_eir}

\input{appendix/online_agg}

\subsection{Running Maximum}

\paragraph{Statement.}
If the sole future consumption of a nonempty prefix is its final maximum, then
$\EIR=b$.

\paragraph{Proof.}
Two prefixes are continuation-equivalent exactly when their maxima agree.
Each $a\in\mathcal A$ is attained by the constant prefix
$(a,\ldots,a)$, so there are at least $q$ classes.  Retaining the maximum is
sufficient, so there are at most $q$ classes.  Hence the class count is $q$
and $\EIR=\log_2q=b$.  Reversing the order proves the corresponding minimum
result.
\hfill$\square$

\subsection{Stepwise Top-k}

At each step, the implemented frontier workload returns the sorted top-$k$
values from the fresh segment and two referenced earlier frontiers. Every
intermediate frontier is scored, while future reference labels are hidden at
the cut. Its exact continuation classes are therefore the greatest fixed-point
bisimulation classes of this labelled transition system. This characterizes
when two retained states are equivalent, but does not provide a closed-form
class count. Exhaustive bisimulation is practical only for reduced alphabets
and windows; the evaluated $m=16$, $k=2$, $q=10^4$ state space is not directly
enumerable. Consequently, no numeric EIR or EIR-based pressure is claimed for
Stepwise Top-$k$ in this study.

\subsection{Store--Recall}

The implemented workload alternates STORE and RECALL steps.  Odd step $u$
stores an independent value $A_u=w_u$.  Even step $u$ recalls one odd STORE id
from $[\max(1,u-m),u-1]$, with the id revealed only when the step arrives.
Let $e(t)$ be the first even step after cut $t$.  If $e(t)$ exceeds the task
horizon, set $R_t=\varnothing$; otherwise define
\[
  R_t=\{s\le t:s\text{ is odd and }\max(1,e(t)-m)\le s\}.
\]

\paragraph{Statement.}
Under full support over eligible future recall ids,
$\EIR^{\rm on}_t=|R_t|\log_2q$.

\paragraph{Proof.}
Only the STORE values with ids in $R_t$ can influence future outputs.  If two
prefixes agree on their labelled values in $R_t$, every future recall of a
prefix value agrees, while later STORE values are common future inputs.
Conversely, if the prefixes differ at any $s\in R_t$, choosing $s$ at step
$e(t)$ distinguishes the continuations immediately.  The independent STORE
values therefore induce exactly $q^{|R_t|}$ continuation classes, and
retaining their labelled values attains the bound.  Thus
$\EIR^{\rm on}_t=\log_2q^{|R_t|}=|R_t|\log_2q$.
\hfill$\square$

\subsection{Full Lookup}

\paragraph{Statement.}
If all $m$ labelled coordinates vary independently over $\mathcal A$ and the
future may query every coordinate, then $\EIR=mb$.

\paragraph{Proof.}
There are $q^m$ independent labelled prefix vectors.  Any two distinct
vectors differ at some coordinate, and querying that coordinate distinguishes
them, so all $q^m$ vectors belong to different continuation classes.
Retaining the labelled vector is sufficient.  Therefore
$\EIR=\log_2q^m=m\log_2q=mb$.
\hfill$\square$

%% file: appendix/computing_eir.tex
\subsection{Exact EIR by finite enumeration}
\label{sec:computing-eir}

Algorithm~\ref{alg:exact-eir} computes exact EIR for finite, nonempty sets of
admissible prefix states and future input sequences, given an evaluator for
the correct outputs of the fixed suffix workload. Here each $u_i$ is an
admissible future input sequence (or query in a one-step suffix), and
$F_t(x,u_i;Z_t)$ is its required answer or output trace; the two are not
interchangeable.

\begin{algorithm}[H]
\caption{Exact EIR by enumeration}
\label{alg:exact-eir}
\begin{algorithmic}[1]
\REQUIRE Prefix states $\mathcal X_t$; future sequences
$\mathcal U_t=(u_1,\ldots,u_m)$ in a fixed order; fixed information $Z_t$;
correct-output evaluator $F_t$ for workload $\mathcal W_t$
\ENSURE $\operatorname{EIR}_t(\mathcal W_t;Z_t)$ in bits
\STATE $\mathcal S\gets\varnothing$ \COMMENT{Set of distinct output signatures}
\FORALL{$x\in\mathcal X_t$}
  \STATE $s_x\gets\bigl(F_t(x,u_1;Z_t),\ldots,F_t(x,u_m;Z_t)\bigr)$
  \STATE $\mathcal S\gets\mathcal S\cup\{s_x\}$
\ENDFOR
\STATE $N_t\gets|\mathcal S|$
\RETURN $\log_2 N_t$
\end{algorithmic}
\end{algorithm}

Two states have the same signature exactly when they are
continuation-equivalent, so $|\mathcal S|$ counts the required equivalence
classes. The procedure requires $|\mathcal X_t|\,|\mathcal U_t|$ evaluations
of $F_t$, in addition to signature comparison and storage, and may be
computationally expensive. When enumeration is impractical, an exact result
requires a continuation-sufficient representation and a matching lower bound
on the number of distinguishable context states.

%% file: appendix/online_agg.tex
\subsection{Stepwise Sum}
\label{sec:online-agg}
The Stepwise Sum task scores all 64 answers, with reward equal to the number of
correct answers divided by 64. At step $u$, the agent receives a new scalar
$w_u$ and references $R_u$, and must output
\[
 A_u=\left(w_u+\sum_{r\in R_u}A_r\right)\bmod q,
 \qquad q=10^4.
\]
References are revealed only when the step arrives. They are selected from
the preceding $m=16$ answers. Consequently, the final-aggregate-only contract
does not characterize this experiment. This is the final-aggregate contrast
highlighted in Section~\ref{sec:eir}: a suffix that asks only for one final sum
can retain an aggregate sufficient statistic, whereas this online suffix may
expose different references at every remaining step and must preserve the
eligible labelled values.

\paragraph{Contract for the online bound.}
Consider a cut after a correct prefix of $t$ steps, before the next references
are revealed. Let $n=\min(t,m)$ and let $x\in\mathbb Z_q^n$ be the ordered
window of eligible answers. We require exact answers for every admissible
future continuation, with no recovery or accessible copy of the prefix in
the model-facing decision channel. The evaluated model receives no tool access
and its decision prompt does not reveal the generator seed. The task container
does contain the seed and generator; this restriction therefore does not
automatically describe a shell-capable agent in that container.
Primitive inputs range freely over $\mathbb Z_q$, independently of the graph.
For $n\ge2$, assume every unordered pair of distinct eligible indices is an
admissible next reference set; for $n=1$, the sole index is referenced.
These assumptions define a full-support task family, not the finite seeded
sample used for evaluation. The bound is a worst-case family-level zero-error
requirement, not the entropy or finite-support EIR of the sampled trials.
Hiding a deterministic seed does not itself establish full support or
independence; the full-support family is the declared theoretical model.
The implementation uses exactly two distinct references
whenever two predecessors exist. If singleton queries were also permitted,
the full-window lower bound would follow immediately from those queries.

\paragraph{Claim: interior cuts.}
If $n\ge3$ and at least two steps remain, the online zero-error requirement is
\[
 \operatorname{EIR}^{\mathrm{on}}_t=n\log_2 q.
\]
Thus, for $16\le t\le62$, the scalar AGG workload has requirement
$16\log_2(10^4)\approx212.603$ bits as a log-cardinality. A jointly coded
fixed-length binary representation of the window requires 213 bits; storing
each scalar in its own integer-width binary slot requires $16\times14=224$
physical bits. The same full-window requirement holds for $n=1$, and for $n=2$
with at least two steps remaining in this $m=16$ workload.

\paragraph{Proof.}
Fix a graph prefix. Every answer vector is attainable: choose primitive
inputs successively as $w_u=A_u-\sum_{r\in R_u}A_r\pmod q$.
The recurrence is triangular, so recursive dependence does not restrict the
full answer-vector domain under the stated input assumption.
Suppose two window states $x,x'$ share a sufficient message. Set
$\delta_i=x'_i-x_i\pmod q$. Equality of the next answer for every admissible
pair requires $\delta_i+\delta_j=0$ for every $i\ne j$.
For $n\ge3$, comparison of overlapping pairs implies all $\delta_i=c$ and
$2c=0$. For even $q$, the only possibilities are $c=0$ and $c=q/2$.
In the latter case, the first future answer is identical for the two states.
At the following step, select that newly produced answer and any surviving
old answer. Their sum differs by $c$, distinguishing the states. Such an old
answer survives because $m=16$. Therefore $c=0$ is necessary, and all $q^n$
states are distinct continuation classes. Retaining the ordered window
achieves this bound: answer the next query, append its result, and discard the
expired oldest value. For $n=2$, equality of the first pair sum followed by a
query combining the new result and either old value also distinguishes every
state. For $n=1$, the next answer distinguishes that value directly.
\hfill$\square$

\paragraph{Terminal boundary.}
If only one step remains, $n\ge3$, and references always contain exactly two
distinct indices, the indistinguishable shifts are precisely the constants
$c$ satisfying $2c=0\pmod q$. There are $\gcd(2,q)$ such shifts, giving
\[
 \operatorname{EIR}^{\mathrm{on}}_t
 =n\log_2q-\log_2\gcd(2,q).
\]
For odd $q$ there is no reduction. For the final scored cut with $q=10^4$,
\[
 \operatorname{EIR}^{\mathrm{on}}_{63}
 =16\log_2(10^4)-1\approx211.603\ \text{bits}.
\]
If singleton reference sets are admissible, this exception disappears.
After the last scored answer, the remaining requirement is zero.

\paragraph{Interpretation of the RQ1 results.}
At full-window
interior cuts, a closed payload of $h$ ordered $q$-ary scalar slots has
log-cardinality capacity $h\log_2q$, and the ideal online pressure is then
$P_t=\EIR_t/C_t=m/h$. We call $m/h$ the interior online pressure; before the window fills,
the corresponding ratio is $n/h$, and the terminal cut has the correction above.
For $h=16,12,8,4$, these pressures are $1,4/3,2,4$.
This identification requires that all prefix-dependent storage and metadata
be accounted for; it does not certify the implemented channel from its name.
For $h<m$, perfect execution cannot be guaranteed over all admissible
continuations. This worst-case statement does not specify an average accuracy
curve or require every sampled trajectory to fail. The observed curves still
measure the combined effects of missing operands, computation, and propagation.

This online contract cannot be characterized by the 13-bit requirement of an
offline final-aggregate task, nor does that offline contract establish an
intrinsic EIR ordering between Stepwise Sum and Store--Recall. Belady has access to future references:
its advantage over an online policy is a hindsight comparison, not by itself
avoidable waste for an agent given the same information as the tested policy.

\paragraph{Implementation verification.}
Source inspection confirms that the
verifier scores all 64 answers, the step interface reveals only current
references, the generator selects two distinct eligible predecessors once two
exist, and the evaluated controlled-retention runs prohibit recovery. The
container contains the deterministic generator and seed, while
the model has no tool access and its decision prompt exposes neither.
The theorem concerns that restricted decision interface under the declared
full-support family. Exact implemented channel capacity still requires accounting
for all prefix-dependent fields and storage. The theorem does not identify the
entropy of the three sampled seeds or guarantee their empirical failure rate.

%% file: appendix/experimental_details.tex
\section{Experimental Configurations and Settings}
\label{sec:app-experimental}

This appendix specifies the configurations behind the controlled results in
Section~\ref{sec:results}. Frozen manifests and compact aggregates are included
in the supplementary archive. Unless stated otherwise, tasks use 64 steps,
four-digit values (arithmetic modulo $10^4$), and graph seeds
$\{1,2,3\}$.

\subsection{Models and decoding}

Table~\ref{tab:model-settings} gives the exact model handles recorded in the
capability-control manifest. Models were resolved through the provider catalog;
the provider-qualified identifier was logged with each run.

\begin{table}[ht]
\centering
\small
\caption{Model and decoding configuration. Temperature zero was sent only
when the endpoint advertised temperature support.}
\label{tab:model-settings}
\begin{tabular}{lll}
\toprule
Paper name & Recorded provider identifier & Decoding \\
\midrule
GPT-5.6 Reasoning & \texttt{dev-gpt-56-reasoning} & provider default$^\dagger$ \\
Claude Opus 4.6 & \texttt{dev-anthropic-claude-opus-4-6} & temperature 0 \\
Claude Haiku 4.5 & \texttt{dev-anthropic-claude-haiku-4-5} & temperature 0 \\
GPT-4.1 Mini & \texttt{dev-gpt-41-mini-04-14} & temperature 0 \\
\bottomrule
\end{tabular}

\vspace{0.3em}
\parbox{0.96\linewidth}{\footnotesize $^\dagger$The reasoning endpoint rejects a
temperature argument, so no temperature was sent. No experiment overrides
\texttt{top\_p}, presence/frequency penalties, or output-token limits; these
therefore follow provider defaults and were not frozen as study parameters.}
\end{table}

The model client permits three provider-layer attempts. Each chat call has a
240-second client timeout; timeout returns an empty reply that the action parser
handles as an invalid decision. Task commands have a 120-second timeout and up
to three attempts, with 5- and 10-second delays between attempts. These retries address
transport or command timeouts, not incorrect model answers.

\subsection{Task runtime and interfaces}

Each generated Harbor task declares a 300-second verifier timeout, a
1,800-second base agent timeout, a 600-second image-build timeout, one CPU,
2,048 MB memory, and 4,096 MB storage. Some batch runners multiply the Harbor
agent timeout: 4 for the Stepwise Maximum restoration replication, 2 for the
Full Lookup adaptation runs, and 8 for the external-memory architecture runs.
The latter also use a six-hour batch-level timeout. Host hardware and provider
region were not recorded and are not used as experimental factors.

Controlled Retention makes a fresh model request at each decision. The request
contains the task rule, current step, the most recent $h$ complete submitted
answer records, and any recovery supplied for that step. No prior model
transcript or scratch state crosses the cut. At most 20 recover/answer decisions
are permitted per step. In no-recovery conditions, \textsc{Read} and
\textsc{Recover} are disabled. Agent-Managed Retention uses the selected
bounded transcript or external-memory implementation. The shared bounded loop
defaults to eight transcript turns and a 600-action ceiling; experiment runners
override these bounds where noted below. The frozen architecture matrix used a
200-turn scoring ceiling; the current 600-action class default is not applied
retroactively to those reported runs.

\subsection{Controlled task grids}

Table~\ref{tab:experiment-grid} summarizes the frozen controlled grids. Here
$m$ is the nominal dependency window, $h$ is retained complete records, $n$ is
task length, and $f_{\rm in}$ is maximum direct fan-in. All model comparisons
reuse byte-identical seeded task instances.

\begin{table*}[t]
\centering
\footnotesize
\setlength{\tabcolsep}{4pt}
\caption{Controlled experiment configurations. Counts are attempted registered
trials unless explicitly called deterministic analyses or probes.}
\label{tab:experiment-grid}
\begin{tabular}{@{}p{2.2cm}p{2.55cm}p{3.15cm}p{2.2cm}p{2.1cm}@{}}
\toprule
Study & Workloads and graph settings & Capacity, recovery, and policies & Models/seeds & Registered count \\
\midrule
Capability control & Five families; $m\in\{4,8,16\}$; $n=64$; $f_{\rm in}=2$ except Running Maximum chain & Full information, $h=m$; no-tool primary plus deterministic-calculator control & Four models; seeds 1--3 & 180 primary; 72 tool-control \\
\addlinespace
No-recomputation pressure & Store--Recall at $m\in\{8,16\}$; Stepwise Sum, Maximum, and Top-$k$ at $m=16$; $n=64$ & Store--Recall $h\in\{2,4,\ldots,16\}$ for $m=16$ and $h\in\{2,4,6,8\}$ for $m=8$; other families $h\in\{4,8,12,16\}$; recovery prohibited & Four models for Store--Recall; three capability-eligible models otherwise; seeds 1--3 & 252 \\
\addlinespace
Store--Recall restoration & Store--Recall, $m=8$, $h=4$, $n=64$ & No injection, exact missing result, or equal-length irrelevant result; recovery otherwise prohibited & Four models; seeds 1--3 & 12 trajectories per arm \\
\addlinespace
Stepwise Maximum restoration & Random-DAG maximum, $m=8$, $h=4$, $n=64$, $f_{\rm in}=2$ & Control, exact unavailable operands, or length-matched sham & Four models; seeds 1--3 & 12 trajectories per arm \\
\addlinespace
Recovery geometry & Sum DAG, $m=8$, $n=64$, $f_{\rm in}\in\{1,2,4\}$; nodes $t\ge32$ & Atomic retrieval versus memoized and naive recursive recomputation & Deterministic; seeds 1--3 & 99 cones/topology \\
\addlinespace
Model recovery probes & Stepwise Sum, Maximum, and Top-$k$ recovery cones & Model selects reads; harness performs arithmetic; charged reads compared with oracle minimum & Evaluated model panel & 88 probes \\
\addlinespace
Full Lookup adaptation & Full Lookup, $m=16$, $n=64$; uniform, Zipf, and midpoint-shift queries & $h\in\{16,8,4\}$; fixed, LRU, and LFU; one atomic read per miss; transcript bound 2 & Four models; seeds 1--3 & 324 \\
\addlinespace
Stepwise Maximum behavior & Random-DAG maximum, $m=8$, $n=64$, $f_{\rm in}=2$ & $h\in\{2,4\}$; floor, Scratch, LRU, Scratch--LRU & Four models; seeds 1--3 & 96 \\
\addlinespace
Architecture comparison & Primary Store--Recall and Stepwise Top-$k$ at $m=8$, $h\in\{2,4\}$; extension covers all six workloads & Strict record-bounded and native-system panels; effective count shown per row & Four models; seeds 1--3 & 24 per architecture--workload row \\
Retrieval Access Budget (registered partial) & Full Lookup, $m=16$, $h\in\{16,8,4\}$ & FIFO eligibility at the adapter boundary; two-turn action context; direct recovery prohibited & Mem0, A-MEM, and Graphiti; four models; seeds 1--3 & 108 planned; 33 persisted \\
\bottomrule
\end{tabular}
\end{table*}
\FloatBarrier

The Full Lookup adaptation grid contains
$4$ models $\times 3$ seeds $\times 3$ capacities $\times 3$ query
distributions $\times 3$ policies $=324$ valid trials. The pressure study uses
the frozen 252-cell list once, with no outcome-adaptive stopping. Reported
architecture tables retain the attempted denominator and exclude runs without
effective outcomes rather than imputing zero.

\subsection{Memory architectures and capacity contracts}

The strict panel uses a common unit of complete submitted answer records.
Floor, FIFO, LRU, LFU, and Belady enforce exactly $h$ records; Belady observes
future references and is an offline, non-causal upper bound. The Mem0 adapter
uses FIFO admission and physical deletion to enforce $h$ stored records at
runtime. Exact-cache agents use a two-turn transcript bound.

Scratch and Scratch--LRU use model-visible text rather than a certified
one-record unit. Forced paging is intentionally unbounded. Letta exposes no
enforceable $h$-record bound; A-MEM exposes no auditable per-record deletion;
Graphiti can expand graph state beyond one record. These systems are therefore
reported in a separate native-system panel and are not interpreted as
equal-capacity policy comparisons. Optional dependency ranges were Letta
\texttt{>=1.12,<2}, Mem0 \texttt{>=2.0.19,<3}, and Graphiti
\texttt{>=0.30.2,<0.31}; A-MEM was pinned to repository revision
\texttt{f303dfc71e07bdc787f4bc135d4cea328ae30e99}.

The \emph{Retrieval Access Budget} provides a common calibrated interface for
systems whose task-state reads pass through the external-memory adapter. Each
backend may retain its native internal state, while a FIFO eligibility set
permits at most $h$ labelled submitted-answer records to cross the retrieval
boundary. The adapter rejects an ineligible identifier before charging or
querying the backend. We fix the action transcript at two turns, do not retain
query outputs, and prohibit direct READ/RECOVER actions. Thus the panel measures
retrieval-interface capacity, not total backend storage or each product's own
eviction policy. We use Full Lookup with $m=16$ and
$h\in\{16,8,4\}$, giving exact pressure $P\in\{1,2,4\}$. Letta is excluded
because its persistent agent owns memory internally and can bypass an
adapter-level retrieval gate. Calibrated Graphiti runs use full episode
ingestion and native episodic BM25/cross-encoder retrieval; the unrestricted
native panel retains its historical exact-episode adapter for reproducibility.
The run was stopped before completion: 33/108 ledger rows were persisted (27
completed A-MEM cells and six Graphiti infrastructure errors). We archive these
rows and their available telemetry for reproducibility but exclude the partial
matrix from headline aggregates and do not impute missing cells as zero.

\subsection{Endpoints, aggregation, and uncertainty}

The verifier reports end-to-end accuracy against ground truth,
self-consistent accuracy against the agent's own upstream answers, answered
fraction, reads, rereads, budget denials, memory writes, and evictions. The
capability screen averages each model--family cell over three seeds and excludes
it from information-loss attribution when full-information mean end-to-end
accuracy is below 0.75. The frozen exclusions are GPT-4.1 Mini on Stepwise Sum,
both maximum workloads, and Stepwise Top-$k$; Store--Recall remains eligible.

Controlled tables report means and population standard deviations over the
stated model--seed cells. Monotonic pressure associations use Spearman rank
correlation. Recovery-geometry figure intervals use seeded nonparametric
bootstrap intervals for the plotted means. The real-trajectory analyses use
their own observational models: task-fixed-effects conditional Poisson models
with run-clustered errors for rereading; task/model-adjusted logistic models for
gap sensitivities; and task-cluster bootstrap standardization for mediation.
The number of observations, interval, and adjustment set are reported beside
each real-trajectory result.

\subsection{Validity, exclusions, and provenance}

A controlled result is excluded when its task manifest is missing, the graph
does not match the registered cell, verifier ground truth is exposed, retrieval
pricing differs from the declared condition, or infrastructure failure prevents
an effective outcome. A bounded run that reaches the action ceiling but leaves a
valid terminal action record is a behavioral outcome and remains scored.

Run manifests record workload, graph fingerprint, operator, seed, model,
retention condition, capacity/context bound, recovery budgets, recovery
mechanism, and source revisions. No study-level override was recorded for
\texttt{top\_p}, output-token limits, penalties, host hardware, or provider
region; these fields must therefore not be treated as controlled variables.

\paragraph{Harness channel audit.}
A falsification audit tested seven potential undeclared channels across 100
synthetic instances and 18{,}068 checks. All checks passed. This audit tests the
implemented interface; the workload contract remains authoritative.

\begin{table}[ht]
\centering
\small
\caption{Controlled-retention interface audit.}
\label{tab:smuggling-audit}
\begin{tabular}{lrr}
\toprule
Channel checked & Checks & Findings \\
\midrule
Payload width & 7{,}777 & 0 \\
Identifier set & 1{,}752 & 0 \\
Occupancy & 1{,}752 & 0 \\
Ordering & 1{,}752 & 0 \\
JSON determinism & 1{,}752 & 0 \\
Message length at saturation & 1{,}531 & 0 \\
Reset isolation & 1{,}752 & 0 \\
\midrule
\textbf{Total} & \textbf{18{,}068} & \textbf{0} \\
\bottomrule
\end{tabular}
\end{table}

%% file: appendix/results_details.tex
\section{EIR Calculations and Detailed Results}
\label{sec:app-results-details}

We first report the full-information capability control and EIR calculations,
then mirror the main-paper order: RQ1 tests prediction and causality, RQ2
measures recovery cost, RQ3 examines adaptation, and RQ4 compares memory
architectures.

\subsection{Full-information capability control}
\label{sec:capability-results}

Table~\ref{tab:capability} reports the no-tool control used to determine which
model--workload cells support information-loss attribution. All three stronger
models clear the pre-registered $0.75$ floor in every tested family.
\texttt{Gpt41Mini} clears Store--Recall but falls below the floor on Stepwise
Sum, both maximum workloads, and Stepwise Top-$k$; those cells are excluded
from the corresponding pressure analyses. In a deterministic-calculator
control, every tested cell reaches 1.000, including all four deficient
\texttt{Gpt41Mini} families and Stepwise Top-$k$ for the stronger models. This
confirms that the exclusions reflect unaided execution capability rather than
information availability.

\input{tables/tab_capability}

The completed error decomposition further isolates the residual capability
gap. Across 576 scored steps per model--family cell, the three stronger models
make no errors on Stepwise Sum or Store--Recall and at most one error on either
maximum workload. Their Stepwise Top-$k$ computation-error rates are 22.4\%
(Haiku), 6.9\% (Opus), and 8.2\% (GPT-5.6), with cascade rates at most 2.3\%.
Thus the Top-$k$ full-information deficit is a computation confound rather than
evidence of unavailable state; the deterministic-calculator control removes it.

\subsection{Exact EIR calculations by workload contract}
\label{sec:app-eir-calculations}

\paragraph{Setup.}
We compute EIR by deriving the continuation-equivalence classes induced by
the complete workload contract---operator, dependency graph, future
consumption, scoring target, and reveal timing, as summarized in
Table~\ref{tab:lacuna-workloads}. Runtime operator variants are instantiated on byte-identical base
graphs generated by the same seed, fan-in, step count, and reference span before
operator selection. The frozen final-query results use the offline contract of
\S\ref{sec:model}; the Stepwise Sum result instead uses the online contract of
the Stepwise Sum result.

\paragraph{Theoretical results.}
The frozen final-output contracts have exact certified EIR values independently
of any Harbor run. For a reference window of $m$ values drawn from an alphabet
of size $q=2^b$:
\begin{itemize}
  \item Running Maximum: $\EIR = b$ bits. At $d=4$, this is
    $\log_2(10^4)\approx13.3$ bits.
  \item Stepwise Maximum: exact EIR is graph- and cut-dependent
    across the three certified $m=16$ graphs; full-window
    cuts range from 175.3 to 194.7 bits, and peak EIR ranges from 190.2 to
    194.7 bits.
  \item Stepwise Sum: $\EIR^{\rm on}=|W_t|b$ when at least two scored
    steps remain, where $W_t$ is the eligible answer window. At
    saturated interior cuts with $m=16$ and $d=4$, this is approximately
    $212.6$ bits.
  \item Store--Recall: peak interior $\EIR^{\rm on}=8b\approx106.3$ bits
    after odd cuts and $7b\approx93.0$ bits after even cuts for $m=16$.
\end{itemize}

\paragraph{Graph identity check.}
The base graph is generated before operator selection and its SHA-256 fingerprint is
identical across runtime operators at matched parameters. The certification
suite verifies this for every combination of seed, $m$, and
$d$ in the experiment grid. Graph identity removes one confound, but does not
make workloads with different reveal or scoring contracts directly comparable.

\paragraph{Matched construction.}
Analytic final-maximum constructions can expose the same $b$-bit extremum after
prefixes of different depth. This statement does not assign $b$ bits to the
Stepwise Maximum runtime.

\paragraph{Enumeration validation.}
The exhaustive enumeration suite enumerates every prefix in $\mathcal A^m$ for
$q\in\{2,3,4\}$ and $m\in\{0,\ldots,4\}$ across all three certified
families. The enumeration and bridge tests pass; class counts match the theorem
formulas exactly at every configuration, and dedicated tests cover the online
Stepwise Sum interior and terminal cases. No discrepancy has been found between the
partition argument and the executable class count.

\newcommand{\rqfourappendix}{%
\subsection{RQ4: How closely do architectures approach the EIR floor?}
\label{sec:results-rq4}

\paragraph{Prediction.}
A concrete memory architecture may realize only part of the EIR floor. The
gap above that floor measures policy
waste. Offline Belady supplies a capacity-matched eviction ceiling, whereas
forced paging supplies an unbounded exact-storage upper bound; FIFO is the
deployed baseline. We predict that task-aligned exact retrieval will close more
of the realization gap than generic recency-based retention.

\paragraph{Results.}
The offline Belady simulation compares record-retention policies on the sampled
graphs without any agent runs
(the detailed Belady table in this appendix).
At $\Pint{=}1.33$ ($h{=}12$, $m{=}16$), FIFO misses on 38.2\% of consuming
steps while hindsight Belady misses on \textbf{zero} for these realized graphs.
This does not establish that $h=12$ suffices for every admissible online
continuation; Belady has future-reference information unavailable to the tested
online policy.
At $\rnom{=}2.0$, the miss-rate gap is 43.4\% (FIFO 64.0\% minus Belady 20.6\%);
at $\rnom{=}4.0$, it is 29.1\%. The gap is largest at moderate nominal load
because Belady can use future-reference knowledge to retain the exact operands that
will be needed, while FIFO blindly evicts them.

This result fixes the interpretation of the later failure curves for the
realized seeded graphs: some FIFO misses are avoidable with hindsight. It does
not convert the hindsight advantage into avoidable waste for an online agent
with the same information as the tested policy.

\emph{LRU: recency-based eviction provides no relief.}
We simulate LRU (evict the least recently \emph{used} answer) on the same graphs.
The result is striking: at $h{=}m{=}16$ (zero-miss, $\Pint=1$ baseline), LRU incurs a
21.2\% miss rate while FIFO has zero.  LRU prematurely evicts answers that have not
been recently referenced but will be needed soon, because sum-task references are
semi-random within the last $m$ steps.
Across nominal-load levels: $\rnom{=}1.33$: LRU 39.7\% (worse than FIFO 38.1\%);
$\rnom{=}2.0$: LRU 61.9\% (marginally better than FIFO 64.0\%);
$\rnom{=}4.0$: LRU 85.7\% (equal to FIFO).
The LRU--Belady gap mirrors the FIFO--Belady gap at every level---recency-of-use
is not a useful signal here. The simulation shows that nominal capacity alone
does not ensure that a recency policy preserves the eligible window. Whether
an online agent exhibits the same gap requires the sufficient-capacity
agent-executed comparison described in the sufficient-capacity follow-up.

\paragraph{Agent-executed sufficient-capacity follow-up.}
Across the three strong models and three seeds, the agent-loop-matched
exact-window policy scores $0.934$ at $h=16$
(Table~\ref{tab:rq2-agg-capacity}). LRU scores only $0.410$ at $h=16$ and
$0.441$ at $h=20$ even though either setting can hold every one of the
16 currently eligible answers. Its first incorrect answer coincides with its
first missing required reference in every GPT-5.6 and Opus run at capacities
16 and 20; Haiku also exhibits some earlier computation errors. Telemetry records
a mean of 13.3 and 5.7 missing-reference steps, respectively. At $h=32$,
LRU has no missing-reference steps and accuracy rises to $0.925$; remaining
errors occur with all operands available and are therefore not attributed to
retention. Correct restoration of every missing submitted answer raises the
capacity-16 LRU score to the exact-window level of $0.934$, whereas matched sham
restoration reaches only $0.444$. These results show that sufficient provisioned capacity
does not guarantee that an online eviction strategy preserves
continuation-required state, and the restoration contrast attributes the
resulting failure to the missing information rather than prompt length.

\paragraph{Theorem-matched Full Lookup architecture comparison.}
All 144 strict equal-capacity trials and all 48 runnable native-local trials
completed with answered fraction 1. Across four models and three seeds, Belady
has the highest strict-panel mean reward: $0.490$ at $h=4$ and $0.668$ at
$h=8$, versus FIFO at $0.396/0.625$, LRU and LFU at
$0.396/0.626$, the fixed floor at $0.397/0.628$, and Mem0 at
$0.417/0.621$. The native panel is not capacity-comparable: unbounded forced
paging reaches $1.000$ at both capacities, while bounded-context scratch reaches
$0.266/0.281$. The subsequent Letta, A-MEM, and Graphiti extension is reported
in Figure~\ref{fig:rq4-agent-architectures} and
Table~\ref{tab:rq5-agent-architectures} as a separate native,
non-capacity-matched panel. Scored counts are shown explicitly; runs without
effective outcomes are excluded rather than assigned zero reward.
Possessing an oracle eviction rule is therefore not by itself sufficient: the
action model must still reveal steps, retrieve state, compute, format, and
submit answers, which is why agent-executed Belady stays far below the
simulated upper bound.

\input{tables/tab_rq2_agg_capacity}

\emph{Cross-family analysis and theory-informed policy.}
Table~\ref{tab:rq5-cross} extends the simulation to all four workload families
with a fifth policy---the
\emph{theory-informed} policy that uses knowledge of the workload structure.

Four results confirm the theory.  \emph{(1) Store--Recall: theory-informed policy achieves
zero miss at $\rnom\!\le\!2$.}  The theory-informed policy for the store/recall
workload always evicts recall-step answers before store-step answers. This
retains all eight active STORE values at $h{=}12$ and $h{=}8$, yielding a miss
rate of exactly $0.000$, identical to Belady in these sufficient-capacity
conditions. At $h{=}4$, theory-informed and Belady miss rates differ
($0.490$ versus $0.073$). \emph{(2) Stepwise Maximum: LRU is worse than FIFO
(efficiency $\le-0.05$ at all pressure levels)}, confirming that recency-of-use
has no predictive value when refs are drawn uniformly from the span window.
\emph{(3) Stepwise Sum and Stepwise Top-$k$: similarly, LRU provides no improvement} over FIFO at any
nominal-load level; efficiency ranges from $-0.23$ to $+0.05$.  \emph{(4) $\rnom^\star$
ordering reflects dependency structure}: Belady first fails (miss $>5\%$) at
Stepwise Sum ($\rnom^\star{=}1.6$) $<$ Stepwise Top-$k$ ($2.0$) $<$ Stepwise Maximum ($2.7$) $<$ Store--Recall ($4.0$),
consistent with the structure of each family's dependency graph and
reference pattern: Store--Recall provides more opportunity for forward-reference
prediction than the random-DAG families.

\input{tables/tab_rq5_cross_family}

\paragraph{Agent-executed architecture matrix.}
The architecture matrix contains 312 cells over four models, Store--Recall
and Stepwise Top-$k$, $h\in\{2,4\}$, and three seeds
(Table~\ref{tab:rq5-agent-architectures}); all 312 have effective outcomes.
The external-system rows for those two workloads are now also complete. The
broader six-workload extension contains all 576/576 effective outcomes
(Table~\ref{tab:rq4-extended-architectures}). It includes Stepwise
Sum, Stepwise Maximum, Full Lookup, and the Running Maximum low-demand control;
every external-system workload row is complete.
Seventy-five cells whose bounded
200-turn run ended before normal verifier finalization are scored from the
append-only action telemetry; this includes 53 max-turn outcomes. These are
behavioral failures rather than infrastructure exclusions, and their
unconditional accuracy remains in the denominator.

Store--Recall separates task-aligned storage from generic agent memory most clearly.
The unbounded forced-paging upper bound is perfect (1.000), Mem0 nearly
matches it (0.996), FIFO reaches 0.811, and Letta reaches 0.370. On Stepwise Top-$k$ the
same ordering weakens but persists: unbounded forced paging reaches 0.501,
the no-recovery floor 0.342, Mem0 0.289, and FIFO 0.283. Letta answers no
Stepwise Top-$k$ cells correctly end-to-end (0.000).

The low agent-executed Belady result (0.060) does not contradict the offline
Belady ceiling. Both Belady and forced paging expose exact, metered
retrieval, but only Belady enforces the stated $h$-record capacity; forced
paging never evicts and is therefore not a same-budget policy. Moreover,
Belady optimizes only eviction. The model still controls the sequential
action loop and arithmetic, so action thrashing and incomplete trajectories
can dominate any cache advantage. RQ4 therefore identifies two distinct
realization gaps: a retention-policy gap, measured by offline simulation,
and an action-execution gap, exposed by end-to-end execution.

\input{tables/tab_rq5_agent_architectures}
\input{tables/tab_rq4_extended_architectures}
}

\subsection{RQ1: Does unavailable required state predict and cause execution errors?}
\label{sec:results-rq1}

\paragraph{Prediction.}
The retention policy's operational predictor is its per-step miss rate, not the EIR/$C$ ratio itself. The predicted empirical signature is a discrete-time first-divergence hazard $P(t^\star = t \mid t^\star \ge t)$ that rises when a required operand first falls outside the context retained by the policy. Same-pressure cells at different lengths should therefore agree, while family ordering should follow graph-structured miss rates rather than intrinsic EIR alone.

\paragraph{Design.}
The primary experiment fixes $B=0$ (no reconstruction), sweeps $h$ from $h=m$ down across
the four instrumented families on matched base graphs, and records $t^\star$ (first
end-to-end-incorrect step). This design does not estimate budget effects because
$B$ remains fixed.

\paragraph{Pilot mechanism check.}
Pilot validation on Stepwise Sum (\texttt{sum}, $m=16$, single seed) establishes the
window-miss mechanism before the full grid.

\emph{Baseline ($h=16$, $\Pint=1.0$).} 64/64 steps correct, zero window misses.
Maximum reference distance is $\le 16 = h$, so every operand remains resident.
No $t^\star$.

\emph{High pressure ($h=4$, $\Pint=4.0$, recovery prohibited).}
First divergence occurs at step $t^\star = 7$.  This coincides exactly with the
first window-miss event: step 7 references step 1 (distance 6 $> h=4$), so the
required operand is no longer resident and the agent cannot compute the correct sum.
A trajectory-level classification separates post-$t^\star$
errors into two types: immediate state-supply errors at miss steps, and later
cascade errors when downstream steps consume already-corrupted intermediates.
The latter are analyzed below in \S\ref{sec:results-cascade}.

The two error modes are cleanly separable from telemetry. \textbf{First
divergence is predicted by the window-miss condition} (ref-distance $>h$).
For Stepwise Sum, the full-support online requirement at interior cuts is $mb$ and
the closed ordered-record channel has capacity $hb$, so its sweep axis is the
online interior pressure $\Pint=m/h$. For the interleaved store/recall Store--Recall
runtime, peak online interior pressure is
$P_{\rm int,max}^{\rm on}=\lceil m/2\rceil/h$ and even cuts use
$\lfloor(m-1)/2\rfloor/h$. Stepwise Maximum and Stepwise Top-$k$ retain $m/h$
only as an operational ratio because their exact continuation classes are
graph-dependent. Per-step miss rate remains the operational policy predictor.

\textbf{Why $\Pint=m/h$ is the applicable pressure.}
The online Stepwise Sum result requires
the ordered window of $m$ eligible residues at full-window interior cuts.
Against an ordered exact-record channel of $h$ scalar slots, the corresponding
pressure is therefore $\Pint=(mb)/(hb)=m/h$. This is a worst-case online
feasibility statement: $h<m$ cannot guarantee exact continuation, but it does
not prescribe average accuracy on the three sampled graphs.

The four-workload pressure grid plus the Store--Recall $m=8$ replication has
completed (252 cells after adding 36 Stepwise Maximum trials).

\paragraph{Full-grid results.}
Table~\ref{tab:rq2} reports end-to-end accuracy by $(h, \text{family})$ for $m=16$,
strong models only ($3{\times}3=9$ trials per cell).
Table~\ref{tab:rq2} reports the accuracy changes across retention ratios.

\input{tables/tab_rq2}
\input{tables/tab_full_lookup_pressure}

Five qualitative findings emerge.

\emph{(1) Monotonic degradation.} Stepwise Sum and Store--Recall show strictly monotone
degradation; Stepwise Top-$k$ is flat at low retention ratio (baseline capability-floor effects)
before declining at higher ratios. The null hypothesis that the retention ratio is non-predictive
is rejected for Stepwise Sum and arbitrary-access ($p < 0.001$ by Spearman rank
correlation across 9 trials at each level).

\emph{(2) Stepwise Maximum degrades without immediate miss/error alignment.}
Its pooled accuracy falls monotonically from $1.000$ at $h=16$ to $0.836$,
$0.770$, and $0.570$ at $h=12,8,4$. Yet the first window miss is not the first
wrong answer in the pressured cells: a missing operand can be irrelevant when
another available operand is already the maximum. This separates state
unavailability from immediate output sensitivity.

\begin{figure}[t]
\centering
\includegraphics[width=0.94\linewidth]{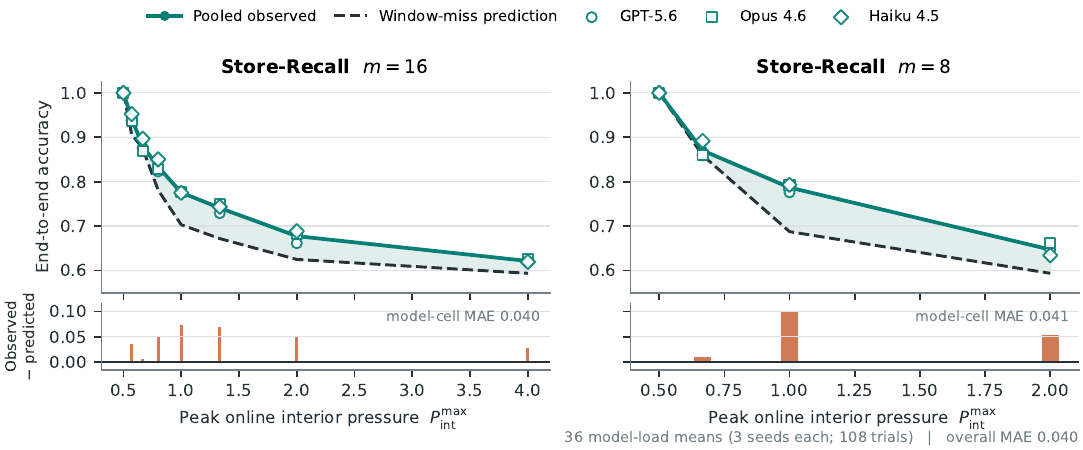}
\caption{\textbf{Store--Recall: theory vs.\ observed.}  Accuracy under the store/recall
arbitrary-access workload compared with the no-lucky-guess prediction
$0.5 + 0.5(1-\text{miss\_rate})$. Open symbols are model means over three
seeds; solid lines are pooled means, dashed lines are predictions, and lower
panels show signed pooled residuals. The close, mostly positive residuals show
that window misses explain the trend without implying exact equality.}
\label{fig:rq2-arb-theory}
\end{figure}

\emph{(3) Store--Recall as a no-cascade baseline.}  The store/recall arbitrary-access workload
isolates the window-miss effect cleanly.  Store steps ($t$ odd, $A_t{=}w_t$) are
always independent; recall steps ($t$ even) reference only store steps and therefore
carry no forward dependency chain. Figure~\ref{fig:rq2-arb-theory} shows that
the window-miss prediction captures the pressure-response shape but is mildly
conservative: the paired model-cell end-to-end residual has mean absolute error
0.040, with observed accuracy generally higher because some missed recalls are
answered correctly.

\emph{(4) Stepwise Top-$k$ capability confound.} Stepwise Top-$k$ (\texttt{frontier}) enters with a sub-unit
baseline (0.861 at $\rnom{=}1$) because frontier sorting imposes a computation demand
that is partially independent of information pressure. After accounting for the per-model
capability floor, the nominal-load effect is still visible (0.861$\to$0.431 from
$\rnom{=}1$ to $\rnom{=}4$), but this design does not separate cascade errors
from computation errors because the capability-controlled tool condition is
not varied across pressure levels.

\paragraph{Comparable online requirement, different failure severity.}
Stepwise Sum and Store--Recall both have online requirements linear in their independently
addressable eligible state. Stepwise Sum nevertheless
degrades much more severely at matched record ratio. The difference lies in
dependency structure: a missed Store--Recall
recall is largely local, whereas an incorrect Stepwise Sum answer can be consumed by
later nodes and propagate through the graph. Thus similar memory requirements
can produce very different failure severity; memory demand alone does not
determine accuracy.

\emph{(5) Causal confirmation: three-arm rescue.}
The pressure curves show that window misses \emph{predict} failure; a counterfactual
intervention shows that the missing state \emph{causes} it.  At every window-miss
recall step (Store--Recall, $m{=}8$, $h{=}4$, recovery prohibited) the decision
prompt is supplied with one of three references (Table~\ref{tab:rescue}): nothing
(\emph{control}), the true missing answer (\emph{oracle}), or an irrelevant value of
matched length drawn from the same task (\emph{sham}).  Oracle injection repairs the
recall completely---miss-step accuracy $1.000$ for all four models---while the matched
sham never repairs it ($0.000$) and control leaves the pressure failure intact
($0.113$ mean, the residual being occasional lucky guesses).  The repair effect
$\Delta_{\rm repair}=\text{oracle}-\text{sham}=+1.000$ is identical across every model
and seed (12 trajectories per arm). Because oracle and sham are matched in length and displayed tokens and
differ only in \emph{correctness}, the effect is attributable to the information
content of the evicted value, not to the presence of an injection or to prompt length.
Restoring exactly the evicted answer---and only that answer---restores correctness,
the strongest causal form of the RQ1 claim.

\input{tables/tab_rescue}
\input{tables/tab_rq1_stepwise_max_rescue}

The Stepwise Maximum replication intervenes on all unavailable operands rather
than only atomic recall steps. Exact restoration yields $1.000$
full-trajectory accuracy for every model, compared with a pooled $0.642$ in
control and $0.714$ under length-matched sham restoration. The
oracle-minus-sham effect is therefore $+0.286$ over the complete trajectory.

\paragraph{Trajectory-level analysis.}
Beyond aggregate accuracy, per-step telemetry reveals that the onset of
failure depends on the workload (Figure~\ref{fig:temporal-alignment}A). First
divergence equals the first window miss in all 27 Stepwise Sum pressure trials.
Exact alignment falls to 0.67 for Store--Recall because a model can occasionally
guess a missing value, delaying the first observed error. For Stepwise Maximum,
the first missing operand does not coincide with first divergence in the
pressure cells: omitting a non-maximal operand can leave the result unchanged.
For Stepwise Top-$k$,
the 0.18 alignment rate reflects the computation confound already visible at
full retention; we therefore do not attribute its first-error timing solely to
information loss.

\noindent\textbf{Summary.} Under FIFO, Stepwise Sum online interior pressure
$\Pint=m/h$ and the realized window-miss rate track accuracy degradation.
Store--Recall quantifies a largely local miss process, and the three-arm counterfactual
rescue confirms that restoring the missing value repairs the failure
($\Delta_{\rm repair}=+1.000$), while a matched irrelevant value does not.
Stepwise Sum and Store--Recall have comparable linear online requirements, but Stepwise Sum
amplifies misses through its dependency graph.

\subsubsection{How do unrepaired information gaps cascade?}
\label{sec:results-cascade}

Once a required value is missed, how far does the resulting error propagate
through downstream dependencies before the trajectory recovers, if it recovers
at all?

\begin{figure}[t]
\centering
\includegraphics[width=0.98\linewidth]{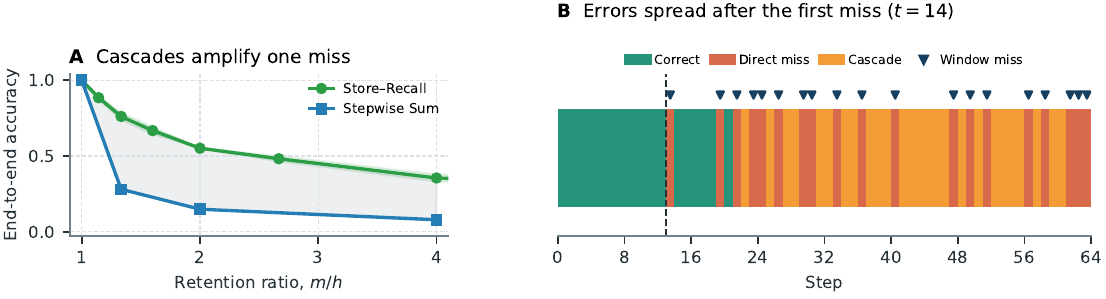}
\caption{\textbf{One missing intermediate result can corrupt many later
steps.} \textbf{A}, Stepwise Sum degrades faster than Store--Recall at matched
$m/h$. Panel A reports accuracy on scored dependency-consuming steps: all
Stepwise Sum steps and RECALL steps only for Store--Recall. Table~\ref{tab:rq2}
instead reports Store--Recall accuracy over both STORE and RECALL steps.
Bands show variation across three seed-pooled model means. \textbf{B}, one
run has its first unavailable dependency at $t{=}14$, followed by 19
state-supply errors, 26 cascade errors, and no computation errors.}
\label{fig:rq2-cascade-gap}
\end{figure}

\paragraph{Cascade amplification in Stepwise Sum.}
Stepwise Sum collapses far below the no-cascade window-miss prediction: at
$\Pint{=}2$ the predicted accuracy from independent-step window misses alone is 0.36,
but the observed value is 0.161: cascades reduce accuracy to approximately
$1/2.2$ of the no-cascade prediction. Cascade errors account
for the remainder: once $A_{t^\star}$ is wrong, every downstream step that references it
inherits the error. Figure~\ref{fig:rq2-cascade-gap}B decomposes this
propagation in a full capability-eligible trajectory.

\paragraph{Dependency amplification is directly visible.}
Store--Recall and Stepwise Sum both have online requirements linear in independently
addressable eligible state, but the interleaved Store--Recall runtime exposes fewer
independent store values at the same nominal $m$. Figure~\ref{fig:rq2-cascade-gap}
therefore compares them at matched operational ratio $m/h$, not exactly matched
EIR. The descriptive accuracy gap grows from zero at $m/h=1$ to
$0.773-0.161=0.612$ at $m/h=2$ and $0.667-0.089=0.578$ at $m/h=4$.
The cross-family subtraction is not itself a causal estimate. Its propagation
interpretation is supported by the within-Stepwise Sum step-level classification:
wrong answers at no-miss steps are self-consistent with already-corrupted
upstream answers.

\begin{figure}[t]
\centering
\textbf{A}\par\vspace{-0.7em}
\includegraphics[width=\linewidth]{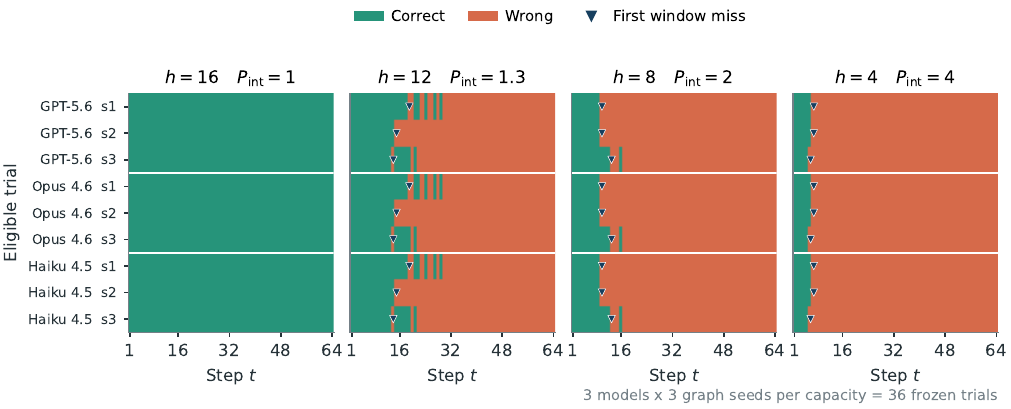}
\vspace{0.2em}
\textbf{B}\par\vspace{-0.7em}
\includegraphics[width=0.72\linewidth]{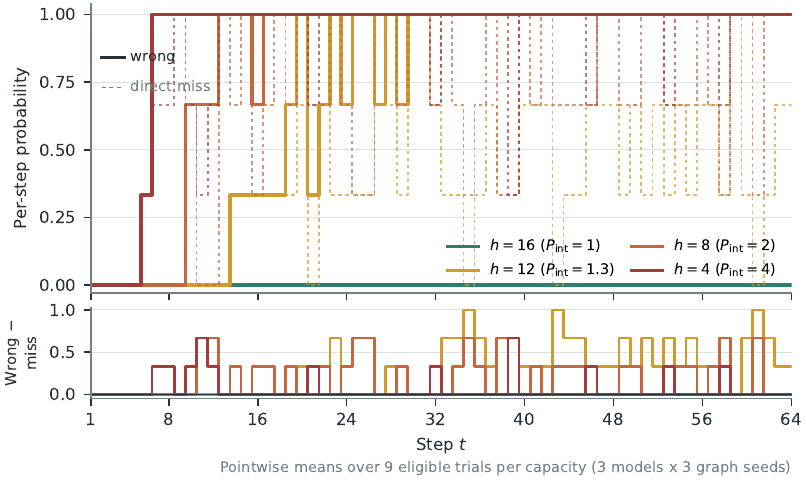}
\caption{\textbf{Stepwise Sum error onset and propagation} ($m{=}16$, recovery
prohibited). \textbf{A:} Each row is one eligible model--seed trial; navy triangles
mark the first window miss. Decreasing $h$ moves the error front earlier.
\textbf{B:} Solid curves show $P(\text{wrong at }t)$ and dashed curves show
$P(\text{direct window miss at }t)$ over the same nine trials per capacity; the
lower panel shows their signed difference. Positive regions mean wrong answers are
more prevalent than direct misses at that step, consistent with downstream cascade
propagation. The 36 ledger-selected trajectories are reproduced from a checksummed
artifact; the capability-control model is excluded.}
\label{fig:rq2-failure-heatmap}
\end{figure}

\paragraph{Errors propagate forward and persist beyond direct misses.}
Figure~\ref{fig:rq2-failure-heatmap} shows all 36 recovered, capability-eligible
Stepwise Sum trajectories (A) and their pointwise aggregate (B). At $h{=}16$ the
heatmap is uniformly green. At lower $h$, an ``error front'' appears at the first
window miss (navy triangle), after which wrong answers dominate. The near-identical
pattern across the three strong models and graph seeds indicates structural onset
rather than a single-model capability failure. Panel B further shows that
wrong-answer probability remains high on later steps without a direct miss. Its
positive wrong-minus-miss gap is the expected signature of corrupted intermediates
propagating beyond the original state-supply error; the abrupt, graph-aligned onset
is inconsistent with a simple gradual-forgetting account.

\paragraph{Error-mode decomposition.}
Figure~\ref{fig:rq2-cascade-gap}B separates direct state-supply errors from
downstream cascades in a complete eligible $h{=}12$ trajectory. The first error
coincides with the first miss at $t^\star{=}14$; over all 64 steps, 19 are direct
state-supply errors and 26 are cascades, with no independent computation errors. Thus the mode
timeline adds information not visible in the correctness-only heatmap.

\noindent\textbf{Summary.} Unrepaired misses do not stay local. In
Stepwise Sum they cascade through the dependency graph, reducing accuracy from
$0.36$ to $0.161$ relative to the no-cascade prediction and producing a persistent post-miss fault
front. Stepwise Sum and Store--Recall have comparable online information
requirements, but Stepwise Sum has more errors because dependency structure
propagates incorrect state. Memory
demand alone therefore does not determine accuracy.

\subsection{RQ2: What determines reconstruction cost?}
\label{sec:results-rq2}

\paragraph{Prediction.}
Recovery cost is determined by the recompute cone, not by pressure.
The predicted law: conditional on a required answer being unavailable, the number of
distinct reads grows with cone depth $D$ and cone size, and is invariant to the
pressure $P$ at which the miss occurred.  A narrow chain and a wide bushy cone of
equal size should have the same unconditional cost but different per-depth costs.

\paragraph{Results.}
The reconstruction-cost denominator is a pure property of the dependency graph: to
recover one evicted answer, an oracle topological recompute must evaluate exactly the
missing node's ancestor cone, with shared ancestors evaluated once.
Table~\ref{tab:rq3-cone} reports this cost across three topologies on matched
\texttt{sum} graphs ($m{=}8$, $n{=}64$, deep nodes $t{\ge}32$, mean over 3 seeds).

\input{tables/tab_rq3_cone}

\begin{figure}[t]
\centering
\includegraphics[width=0.43\linewidth]{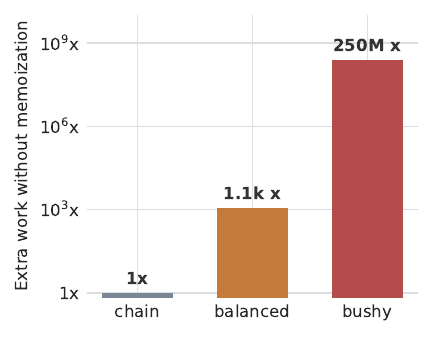}\hfill
\includegraphics[width=0.53\linewidth]{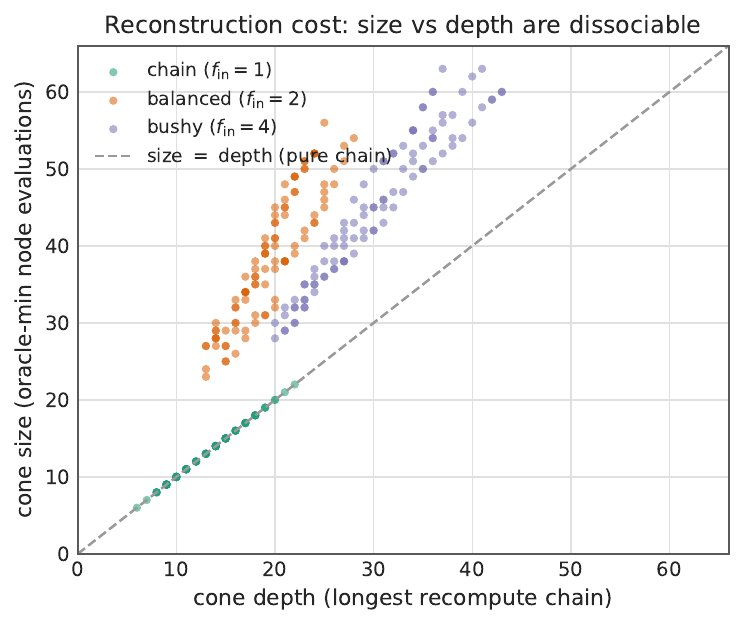}
\caption{\textbf{Dependency geometry determines recovery work and latency.}
\textbf{Left:} Extra work without memoization relative to the cone-size minimum
(log scale, \texttt{sum}, deep nodes $t{\ge}32$). Repeated shared-ancestor traversal
grows from $1\times$ for chains to $2.5{\times}10^{8}$ for bushy DAGs.
\textbf{Right:} Cone size (total recompute volume) versus cone depth (longest
sequential chain) for every non-leaf node. The chain lies on the diagonal, while
balanced and bushy DAGs lie above it, dissociating volume from latency.}
\label{fig:rq3-cost}
\label{fig:rq3-geometry}
\end{figure}

Three results confirm the cost-cone law.
\emph{(1) Atomic recovery is geometry-invariant:} retrieving a stored answer costs one
step regardless of topology---the reference point.
\emph{(2) Recursive-recompute cost is the cone size:} $13.0$ (chain), $38.5$ (balanced),
$45.3$ (bushy) node evaluations, growing with graph density rather than with the
pressure at which the miss occurred.
\emph{(3) The memoization benefit is enormous and geometry-dependent:} without
shared-ancestor reuse the cost is the naive path-recount, which explodes from $13$
(chain, no sharing) to $4.4{\times}10^{4}$ (balanced) to $1.1{\times}10^{10}$ (bushy)---a
$2.5{\times}10^{8}$ memoization speedup in the bushy case (Figure~\ref{fig:rq3-cost}, left).  The
chain is pure depth (cone size equals depth, zero sharing); the bushy DAG is dominated by
shared ancestors.

The size and depth axes are dissociable (Figure~\ref{fig:rq3-cost}, right), so the factorial
contrasts hold on real graphs:
a matched-size pair ($|\text{cone}|{=}56$) differs in depth ($25$ vs $40$)---equal
reconstruction cost at different chain length---while a matched-depth pair (depth $21$)
differs in cone size ($21$ vs $48$)---equal longest-chain latency at different recompute
volume. Cost tracks cone \emph{size} (total evaluations); depth bounds the
critical path under parallel execution, while fully sequential latency tracks
total work. The two are decoupled, exactly as the
cost-cone prediction requires.

\begin{figure}[p]
\centering
\includegraphics[width=0.90\linewidth]{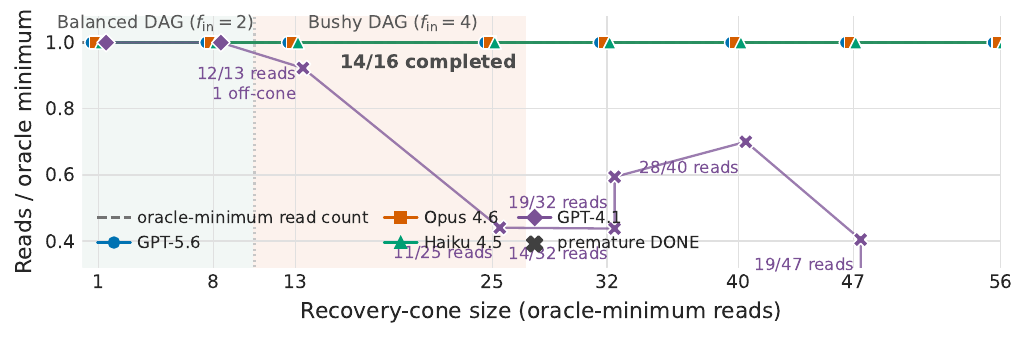}
\caption{\textbf{When recovery succeeds it is read-optimal; the observed
failures stop too soon.} Read overhead is reads divided by the oracle minimum
(1.0 is optimal). All 14 completed probes equal 1.0; crosses mark the two
GPT-4.1 probes that issue \textsc{Done} before reading the full recovery cone.
All 16 probes in the original Stepwise Sum subset are shown.}
\label{fig:rq3-trace}
\end{figure}

\emph{Agent-executed recovery.}  We close the loop with reconstruction probes that
isolates traversal policy from arithmetic: the model is given one evicted target and issues
\texttt{READ} commands to walk the cone, while the harness computes each node's answer once
its references are known (no arithmetic is asked of the model), and we count reads against
the oracle minimum (Figure~\ref{fig:rq3-trace},
Tables~\ref{tab:rq3-amp} and~\ref{tab:rq2-recovery-workloads}). Across 88 probes,
32/40 Stepwise Sum, 18/24 Stepwise Maximum, and 19/24 Stepwise Top-$k$ probes
complete. Every successful probe has amplification exactly $1.000$ with no
redundant or off-cone reads: capable models read each cone node once, following
the memoized recursive policy the cost-cone predicts. The remaining 19 probes
terminate before reconstructing the target rather than exceeding the oracle
read count. Read-order traces for the original Stepwise Sum subset
(Figure~\ref{fig:rq3-trace}) show agents descending to the resident frontier;
successful runs complete the bottom-up recompute, whereas premature runs stop
partway. The risk at scale is incomplete reconstruction, not inefficient
successful traversal.

\paragraph{Cone growth over the trajectory.}
Typical cone size increases with step $t$ because later nodes can accumulate more
ancestors; balanced and bushy trends separate sharply from the chain after step 20.
Individual nodes still vary with graph structure. Across the sampled nodes, the chain
distribution is narrow (max 22, median 9), whereas the bushy distribution is broad
(max 63, median 30), with 50\% of nodes requiring more than 30 ancestor evaluations.
These trajectory-position and distribution summaries are consistent with the joint
size--depth geometry in Figure~\ref{fig:rq3-cost}, rather than independent evidence.

\input{tables/tab_rq3_amplification}
\input{tables/tab_rq2_recovery_workloads}

\subsection{RQ3: How does retention behavior change under information pressure?}
\label{sec:results-rq3}

\paragraph{Prediction.}
Agent-Managed Retention introduces error modes beyond insufficient capacity: repeated
eviction--retrieval cycles on the same item without task progress (thrashing), loss
of already-retrieved state before use (regret), and unacknowledged continuation
after a required value becomes unavailable. A non-monotone accuracy valley is
predicted as nominal load approaches and then exceeds capacity---the agent begins
over-retrieving before continuing without the required value.

\paragraph{Results.}
The theorem-matched Full Lookup sweep completed 324/324 valid trials. At
$h=16$, fixed, adaptive LRU, and adaptive LFU all achieve hit rate 1 with zero
recovery reads. Zipf locality exposes the policy gap: at $h=4$, LFU/LRU/fixed
hit rates are $0.493/0.438/0.111$; at $h=8$ they are
$0.681/0.674/0.174$. Uniform streams show no adaptive advantage. Because each
miss receives one atomic recovery read, mean query accuracy remains at least
$0.998$ in every cell; hit rate and miss regret, not accuracy, identify
retention adaptation in this experiment.

\paragraph{Recovery-enabled Stepwise Maximum comparison.}
A recovery-enabled experiment provides a comparison for the Stepwise Maximum
contract ($\operatorname{max}$, fan-in 2, $m=16$, 64 sequential steps) at
$m/h\in\{1,2,4,8\}$ with three seeds. Unlike the RQ1 grid, this single-agent
protocol permits recovery and scores the final 32 steps. Mean end-to-end
accuracy is $1.000$, $0.948$, $0.938$, and $0.854$ as pressure increases, while
mean recovery reads rise from $9.3$ to $98.3$, $184.0$, and $322.3$. Thus the
run varies Stepwise Maximum nominal load, but its principal signal is
adaptive retrieval that preserves accuracy under pressure; it is not a
no-recovery retention curve and is distinct from the completed RQ1
no-recovery sweep. These results are reported without a model-specific
breakdown because the retained aggregate identifies runs but not model names.

Agent-Managed Retention sweeps produced effective outcomes for 516/526 cells (98.1\%, covering all
four models with deep results for Opus, Gpt41Mini, and Gpt56Reasoning; Haiku at 93.3\%
due to extreme slowness of cap=16 cells at 100k+ tokens per turn).

The Stepwise Maximum behavior panel has all 96/96 effective outcomes. Every
model contributes six outcomes to each of the floor, Scratch, LRU, and
Scratch--LRU rows in Table~\ref{tab:rq3-stepwise-max-behaviors}. Pooled accuracy
is $0.654$ for the no-recovery floor, $0.223$ for Scratch, $0.222$ for LRU, and
$0.072$ for Scratch--LRU. Retrieval is predominantly just-in-time, but the
additional actions do not improve accuracy on this random-DAG extremum workload.

\input{tables/tab_rq3_stepwise_max_behaviors}

\paragraph{Headline.}
Model-driven retention (Strategic Scratch) outperforms do-nothing floor control
in capable models (Opus, Gpt56Reasoning) when paired with \emph{model capability}.
Storage medium dominates: K-V LRU beats text Scratch-LRU by $\approx 0.29$ across all models.
Smaller models (Haiku) struggle with model-driven eviction, reverting to LRU or
floor-like performance.

\paragraph{2$\times$2 confound isolation.}
\emph{Eviction policy} (model-driven vs.~LRU): Opus achieves Scratch acc=0.762
vs.~LRU acc=0.664 ($+$0.098, favour scratch). Gpt56Reasoning matches (Scratch
0.750 vs.~LRU 0.650, $+$0.100). Haiku inverts (Scratch 0.438 vs.~LRU 0.616,
$-$0.178 penalty), showing capability floor where model-driven eviction fails.
\emph{Storage medium} (text vs.~K-V): LRU beats Scratch-LRU by +0.291 (Haiku),
+0.259 (Opus), +0.290 (Gpt41Mini), +0.330 (Gpt56). K-V store advantage is
consistent and large, independent of model.

\paragraph{Capacity scaling.}
All models show graceful degradation as $h$ decreases from 16 to 2. Opus's
scratch baseline maintains high accuracy even at $h=2$ (0.487), mirroring oracle
Belady (0.539). Gpt41Mini reaches Belady-ceiling at $h=8$ and beyond (both 1.0).
Haiku collapses at $h \le 4$ for model-driven scratch (0.0--0.226), recovering
only at $h=16$ (0.953 on 3 seeds).

\paragraph{Valley signature.}
Negative valley deltas emerge in Haiku's scratch baseline: valley$_{\text{min}}$
$= -0.432$ at cap=4. Opus shows local minima but smaller: -0.078 at cap=16.
Gpt56 valley$_{\text{min}} = -0.688$ at Scratch-LRU cap=16, indicating
over-retrieval trap at extreme pressure. Thrash episodes (runs of 3+ non-answer
turns between boundaries) accompany negative valleys, peaking at 12+ per trial
for LRU and Opus at low capacity, declining as pressure eases.

\paragraph{Behavior and mechanism.}
Just-in-time retrieval rate: capable models exhibit 1.0 (Gpt41Mini, Gpt56 scratch
at high cap), while smaller models probe more proactively (Haiku 0.471 at cap=2,
Opus 0.727). Redundant retrieval (same ID reread in one step) clusters in Gpt41Mini
and Gpt56 at cap=2 (0.3--0.4), signalling poor memory efficiency under pressure.
Prefetch precision (proactive reads used within K=4 steps) remains low ($< 0.05$
for most), suggesting model-driven anticipation lacks substrate-level foresight.
Write-through rate (scratch writes after correct answer) varies: Opus peaks at
1.0 (cap=16 scratch), Gpt41Mini 0.4 (cap=2 scratch-lru), Haiku near 0.0 under
model-driven eviction---consistent with reduced planning in small models.

\paragraph{Summary.}
RQ3 reveals a ``model-capability floor'' for Agent-Managed Retention. Opus and
Gpt56Reasoning harness model-driven eviction to beat both floor control and LRU,
approaching oracle Belady under low pressure. Gpt41Mini matches LRU but not
scratch. Haiku fails at model-driven eviction, suggesting that LLM-driven memory
decisions require sufficient model depth. Storage medium (K-V vs.~text) dominates
and benefits all models equally, making it a robust architectural choice
independent of retrieval strategy.

\subsection{RQ4: How closely do architectures approach the EIR floor?}
\label{sec:results-rq4}

\paragraph{Prediction.}
A concrete memory architecture may realize only part of the EIR floor. The
gap above that floor measures policy
waste. Offline Belady supplies a capacity-matched eviction ceiling, whereas
forced paging supplies an unbounded exact-storage upper bound; FIFO is the
deployed baseline. We predict that task-aligned exact retrieval will close more
of the realization gap than generic recency-based retention.

\paragraph{Results.}
The offline Belady simulation compares record-retention policies on the sampled
graphs without any agent runs
(the detailed Belady table in this appendix).
At $\Pint{=}1.33$ ($h{=}12$, $m{=}16$), FIFO misses on 38.2\% of consuming
steps while hindsight Belady misses on \textbf{zero} for these realized graphs.
This does not establish that $h=12$ suffices for every admissible online
continuation; Belady has future-reference information unavailable to the tested
online policy.
At $\rnom{=}2.0$, the miss-rate gap is 43.4\% (FIFO 64.0\% minus Belady 20.6\%);
at $\rnom{=}4.0$, it is 29.1\%. The gap is largest at moderate nominal load
because Belady can use future-reference knowledge to retain the exact operands that
will be needed, while FIFO blindly evicts them.

This result fixes the interpretation of the later failure curves for the
realized seeded graphs: some FIFO misses are avoidable with hindsight. It does
not convert the hindsight advantage into avoidable waste for an online agent
with the same information as the tested policy.

\emph{LRU: recency-based eviction provides no relief.}
We simulate LRU (evict the least recently \emph{used} answer) on the same graphs.
The result is striking: at $h{=}m{=}16$ (zero-miss, $\Pint=1$ baseline), LRU incurs a
21.2\% miss rate while FIFO has zero.  LRU prematurely evicts answers that have not
been recently referenced but will be needed soon, because sum-task references are
semi-random within the last $m$ steps.
Across nominal-load levels: $\rnom{=}1.33$: LRU 39.7\% (worse than FIFO 38.1\%);
$\rnom{=}2.0$: LRU 61.9\% (marginally better than FIFO 64.0\%);
$\rnom{=}4.0$: LRU 85.7\% (equal to FIFO).
The LRU--Belady gap mirrors the FIFO--Belady gap at every level---recency-of-use
is not a useful signal here. The simulation shows that nominal capacity alone
does not ensure that a recency policy preserves the eligible window. Whether
an online agent exhibits the same gap requires the sufficient-capacity
agent-executed comparison described in the sufficient-capacity follow-up.

\paragraph{Agent-executed sufficient-capacity follow-up.}
Across the three strong models and three seeds, the agent-loop-matched
exact-window policy scores $0.934$ at $h=16$
(Table~\ref{tab:rq2-agg-capacity}). LRU scores only $0.410$ at $h=16$ and
$0.441$ at $h=20$ even though either setting can hold every one of the
16 currently eligible answers. Its first incorrect answer coincides with its
first missing required reference in every GPT-5.6 and Opus run at capacities
16 and 20; Haiku also exhibits some earlier computation errors. Telemetry records
a mean of 13.3 and 5.7 missing-reference steps, respectively. At $h=32$,
LRU has no missing-reference steps and accuracy rises to $0.925$; remaining
errors occur with all operands available and are therefore not attributed to
retention. Correct restoration of every missing submitted answer raises the
capacity-16 LRU score to the exact-window level of $0.934$, whereas matched sham
restoration reaches only $0.444$. These results show that sufficient provisioned capacity
does not guarantee that an online eviction strategy preserves
continuation-required state, and the restoration contrast attributes the
resulting failure to the missing information rather than prompt length.

\begin{figure}[p]
\centering
\includegraphics[width=0.96\linewidth]{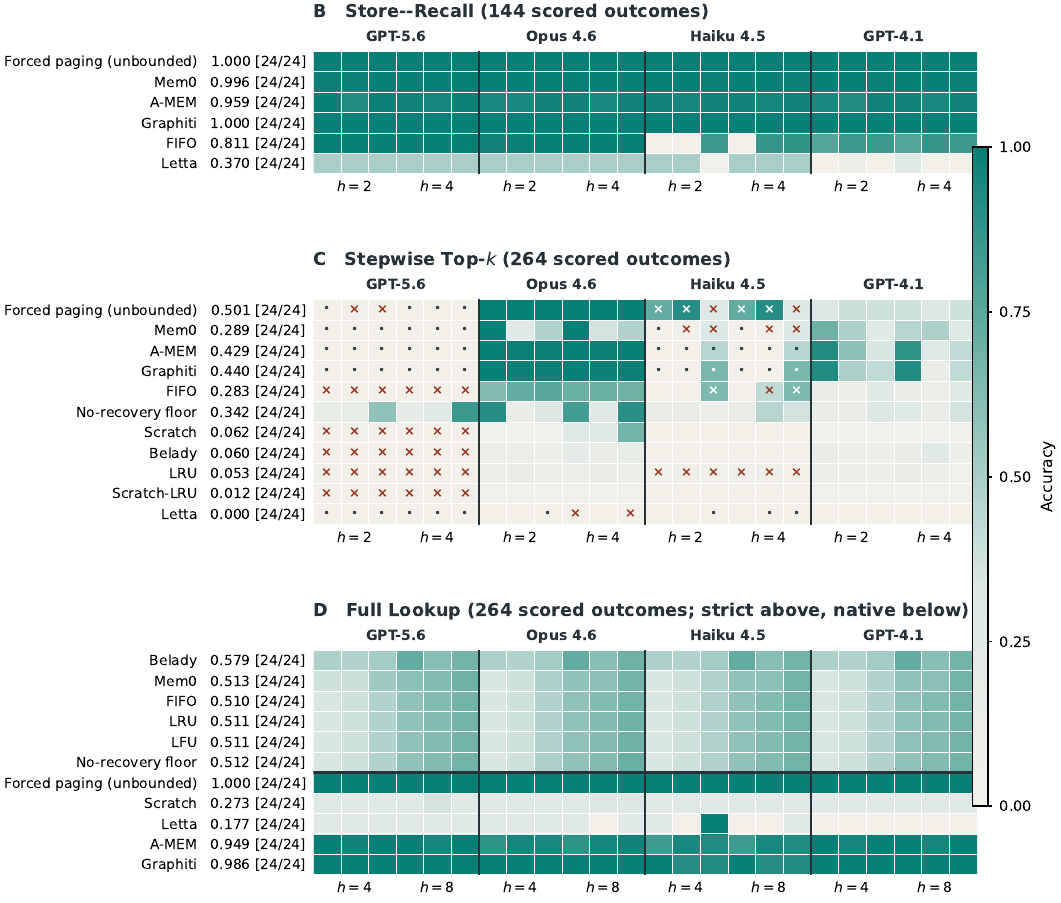}
\caption{\textbf{Architecture outcomes depend on both storage and execution.}
Agent-executed comparisons on Store--Recall, Stepwise Top-$k$, and Full Lookup.
Row labels report means and effective scored counts; crosses mark turn-limit
outcomes. Equal-capacity controls and native systems are shown separately.}
\label{fig:rq4-agent-architectures}
\end{figure}

\paragraph{Theorem-matched Full Lookup architecture comparison.}
All 144 strict equal-capacity trials and all 48 runnable native-local trials
completed with answered fraction 1. Across four models and three seeds, Belady
has the highest strict-panel mean reward: $0.490$ at $h=4$ and $0.668$ at
$h=8$, versus FIFO at $0.396/0.625$, LRU and LFU at
$0.396/0.626$, the fixed floor at $0.397/0.628$, and Mem0 at
$0.417/0.621$. The native panel is not capacity-comparable: unbounded forced
paging reaches $1.000$ at both capacities, while bounded-context scratch reaches
$0.266/0.281$. The subsequent Letta, A-MEM, and Graphiti extension is reported
in Figure~\ref{fig:rq4-agent-architectures} and
Table~\ref{tab:rq5-agent-architectures} as a separate native,
non-capacity-matched panel. Scored counts are shown explicitly; runs without
effective outcomes are excluded rather than assigned zero reward.
Possessing an oracle eviction rule is therefore not by itself sufficient: the
action model must still reveal steps, retrieve state, compute, format, and
submit answers, which is why agent-executed Belady stays far below the
simulated upper bound.

\input{tables/tab_rq2_agg_capacity}

\emph{Cross-family analysis and theory-informed policy.}
Table~\ref{tab:rq5-cross} extends the simulation to all four workload families
with a fifth policy---the
\emph{theory-informed} policy that uses knowledge of the workload structure.

Four results confirm the theory.  \emph{(1) Store--Recall: theory-informed policy achieves
zero miss at $\rnom\!\le\!2$.}  The theory-informed policy for the store/recall
workload always evicts recall-step answers before store-step answers. This
retains all eight active STORE values at $h{=}12$ and $h{=}8$, yielding a miss
rate of exactly $0.000$, identical to Belady in these sufficient-capacity
conditions. At $h{=}4$, theory-informed and Belady miss rates differ
($0.490$ versus $0.073$). \emph{(2) Stepwise Maximum: LRU is worse than FIFO
(efficiency $\le-0.05$ at all pressure levels)}, confirming that recency-of-use
has no predictive value when refs are drawn uniformly from the span window.
\emph{(3) Stepwise Sum and Stepwise Top-$k$: similarly, LRU provides no improvement} over FIFO at any
nominal-load level; efficiency ranges from $-0.23$ to $+0.05$.  \emph{(4) $\rnom^\star$
ordering reflects dependency structure}: Belady first fails (miss $>5\%$) at
Stepwise Sum ($\rnom^\star{=}1.6$) $<$ Stepwise Top-$k$ ($2.0$) $<$ Stepwise Maximum ($2.7$) $<$ Store--Recall ($4.0$),
consistent with the structure of each family's dependency graph and
reference pattern: Store--Recall provides more opportunity for forward-reference
prediction than the random-DAG families.

\input{tables/tab_rq5_cross_family}

\paragraph{Agent-executed architecture matrix.}
The architecture matrix contains 312 cells over four models, Store--Recall
and Stepwise Top-$k$, $h\in\{2,4\}$, and three seeds
(Table~\ref{tab:rq5-agent-architectures}); all 312 have effective outcomes.
The external-system rows for those two workloads are now also complete. The
broader six-workload extension contains all 576/576 effective outcomes
(Table~\ref{tab:rq4-extended-architectures}). It includes Stepwise
Sum, Stepwise Maximum, Full Lookup, and the Running Maximum low-demand control;
every external-system workload row is complete.
Seventy-five cells whose bounded
200-turn run ended before normal verifier finalization are scored from the
append-only action telemetry; this includes 53 max-turn outcomes. These are
behavioral failures rather than infrastructure exclusions, and their
unconditional accuracy remains in the denominator.

Store--Recall separates task-aligned storage from generic agent memory most clearly.
The unbounded forced-paging upper bound is perfect (1.000), Mem0 nearly
matches it (0.996), FIFO reaches 0.811, and Letta reaches 0.370. On Stepwise Top-$k$ the
same ordering weakens but persists: unbounded forced paging reaches 0.501,
the no-recovery floor 0.342, Mem0 0.289, and FIFO 0.283. Letta answers no
Stepwise Top-$k$ cells correctly end-to-end (0.000).

The low agent-executed Belady result (0.060) does not contradict the offline
Belady ceiling. Both Belady and forced paging expose exact, metered
retrieval, but only Belady enforces the stated $h$-record capacity; forced
paging never evicts and is therefore not a same-budget policy. Moreover,
Belady optimizes only eviction. The model still controls the sequential
action loop and arithmetic, so action thrashing and incomplete trajectories
can dominate any cache advantage. RQ4 therefore identifies two distinct
realization gaps: a retention-policy gap, measured by offline simulation,
and an action-execution gap, exposed by end-to-end execution.

\input{tables/tab_rq5_agent_architectures}
\input{tables/tab_rq4_extended_architectures}

\begin{figure}[t]
\centering
\includegraphics[width=0.90\linewidth]{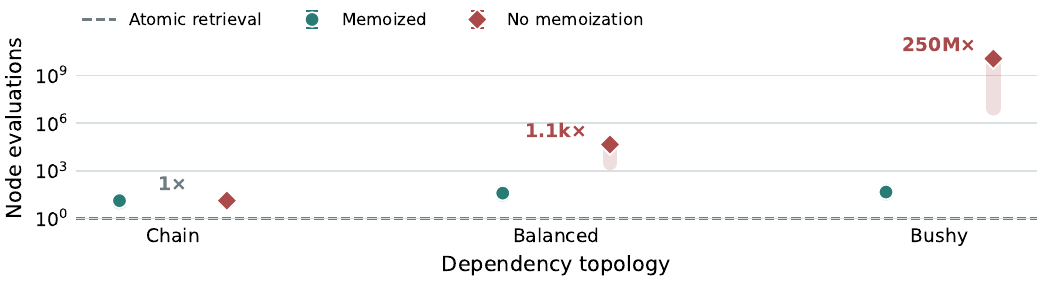}
\caption{\textbf{Reusing computed results avoids repeated recovery work.}
Reconstruction costs with and without reuse of shared ancestors, across
99 cones per graph type. Labels give mean cost ratios; the dashed line
marks one-read retrieval. Pale bars show interquartile ranges, and error
bars show bootstrap 95\% confidence intervals.}
\label{fig:results-recovery}
\end{figure}

\begin{figure}[p]
\centering
\includegraphics[width=0.96\linewidth]{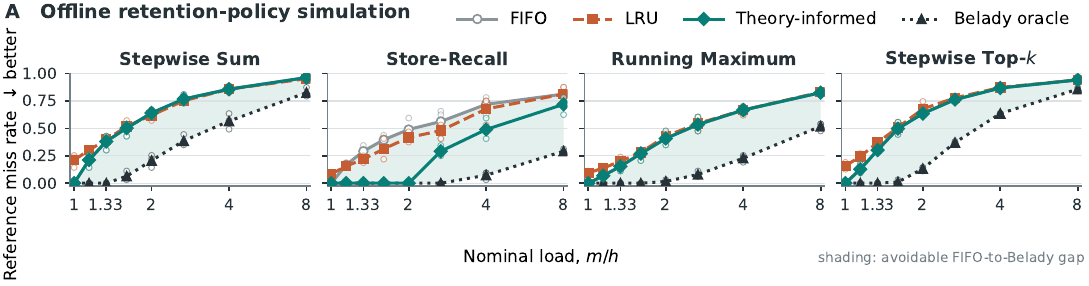}
\caption{\textbf{Future-aware retention can reduce misses.} Simulated policy
miss rates across workloads; Belady uses future requests. Table~\ref{tab:rq5-cross}
reports the underlying values.}
\label{fig:results-policy-cross-family}
\end{figure}

%% file: tables/tab_capability.tex
\begin{table}[H]
\centering
\small
\caption{\textbf{Full-information capability control.} End-to-end accuracy at
$h=m$, averaged over $m\in\{4,8,16\}$ and three seeds (nine trials per entry).
Shaded entries fall below the pre-registered $0.75$ inclusion threshold.}
\label{tab:capability}
\resizebox{\linewidth}{!}{%
\begin{tabular}{lccccc}
\toprule
Model & Stepwise Sum & Store--Recall & Running Max. & Stepwise Max. & Stepwise Top-$k$ \\
\midrule
\texttt{Gpt56Reasoning} & 1.000 & 1.000 & 1.000 & 1.000 & 0.908 \\
\texttt{ClaudeOpus46}   & 1.000 & 1.000 & 0.998 & 1.000 & 0.918 \\
\texttt{ClaudeHaiku45}  & 1.000 & 1.000 & 1.000 & 1.000 & 0.754 \\
\texttt{Gpt41Mini}      & \cellcolor{red!12}0.040 & 1.000 & \cellcolor{red!12}0.474 & \cellcolor{red!12}0.547 & \cellcolor{red!12}0.221 \\
\bottomrule
\end{tabular}%
}
\end{table}

%% file: tables/tab_rq2_agg_capacity.tex
\begin{table}[t]
\centering\small
\caption{\textbf{RQ1 sufficient-capacity follow-up.} Stepwise Sum with the three strong
models (mean $\pm$ population std over $3$ models $\times$ $3$ seeds).
The exact-window and LRU agents use the same managed-memory action loop.
Missing steps count steps with at least one required answer unavailable. Exact-window
retention succeeds at the $h{=}16$ requirement; LRU does not until overprovisioned to
$h{=}32$, and correct restoration---unlike sham restoration---closes the gap.}
\label{tab:rq2-agg-capacity}
\begin{tabular}{lrrrr}
\toprule
Policy & $h$ & \shortstack{End-to-end\\accuracy} & \shortstack{Self-consistent\\accuracy} & \shortstack{Missing-reference\\steps} \\
\midrule
Exact window & 16 & $0.934\pm0.187$ & $0.997\pm0.010$ & $0.0$ \\
LRU & 16 & $0.410\pm0.126$ & $0.788\pm0.044$ & $13.3$ \\
LRU & 20 & $0.441\pm0.145$ & $0.903\pm0.014$ & $5.7$ \\
LRU & 32 & $0.925\pm0.185$ & $0.995\pm0.010$ & $0.0$ \\
\midrule
LRU + correct restoration & 16 & $0.934\pm0.187$ & $0.997\pm0.010$ & $13.3$ \\
LRU + sham restoration & 16 & $0.444\pm0.063$ & $0.790\pm0.049$ & $13.3$ \\
\bottomrule
\end{tabular}
\end{table}

%% file: tables/tab_rq5_cross_family.tex
\begin{table}[t]
\centering\small
\caption{\textbf{Knowing future use matters more than recency.} Miss rate
(lower is better) by retention policy and pressure, averaged over three seeds
with $m{=}16$. Belady is the future-aware optimum. ``LRU gap closed'' is the
fraction of the FIFO-to-Belady gap removed by LRU: $1$ matches Belady, $0$
matches FIFO, and negative values are worse than FIFO. The theory-informed
policy matches Belady for Store--Recall at the two sufficient-capacity settings, whereas
LRU closes little or none of the gap across workloads.}
\label{tab:rq5-cross}
\begin{tabular}{ll rrrr r}
\toprule
Workload & Pressure $m/h$ & FIFO & LRU & Theory-informed & Belady & \shortstack{LRU gap\\closed} \\
\midrule
  Stepwise Sum & 1.33 & 0.381 & 0.397 & 0.381 & \textbf{0.000} & -0.04 \\
   & 2.00 & 0.640 & 0.619 & 0.640 & \textbf{0.206} & +0.05 \\
   & 4.00 & 0.857 & 0.857 & 0.857 & \textbf{0.566} & +0.00 \\
  \midrule
  Store--Recall & 1.33 & 0.292 & 0.219 & 0.000 & \textbf{0.000} & +0.25 \\
   & 2.00 & 0.490 & 0.417 & 0.000 & \textbf{0.000} & +0.15 \\
   & 4.00 & 0.719 & 0.677 & 0.490 & \textbf{0.073} & +0.06 \\
  \midrule
  Stepwise Maximum & 1.33 & 0.153 & 0.201 & 0.153 & \textbf{0.000} & -0.31 \\
   & 2.00 & 0.407 & 0.429 & 0.407 & \textbf{0.016} & -0.05 \\
   & 4.00 & 0.667 & 0.667 & 0.667 & \textbf{0.228} & +0.00 \\
  \midrule
  Stepwise Top-$k$ & 1.33 & 0.302 & 0.370 & 0.302 & \textbf{0.000} & -0.23 \\
   & 2.00 & 0.635 & 0.677 & 0.635 & \textbf{0.132} & -0.08 \\
   & 4.00 & 0.868 & 0.873 & 0.868 & \textbf{0.635} & -0.02 \\
  \midrule
\bottomrule
\end{tabular}
\end{table}

%% file: tables/tab_rq5_agent_architectures.tex
\begin{table}[t]
\centering\small
\caption{\textbf{RQ4 architecture comparison.} Agent-executed end-to-end accuracy at $m{=}8$, pooled over $h\in\{2,4\}$, four models, and three seeds. Every architecture row contains all 24 registered model--capacity--seed outcomes.}
\label{tab:rq5-agent-architectures}
\begin{tabular}{llrrr}
\toprule
Workload & Architecture & Scored & Accuracy & Max-turn \\
\midrule
Store--Recall & Forced paging (unbounded) & 24/24 & 1.000 & 0 \\
 & Mem0 & 24/24 & 0.996 & 0 \\
 & A-MEM & 24/24 & 0.959 & 0 \\
 & Graphiti & 24/24 & 1.000 & 0 \\
 & FIFO & 24/24 & 0.811 & 0 \\
 & Letta & 24/24 & 0.370 & 0 \\
\midrule
Stepwise Top-$k$ & Forced paging (unbounded) & 24/24 & 0.501 & 8 \\
 & Mem0 & 24/24 & 0.289 & 4 \\
 & A-MEM & 24/24 & 0.429 & 0 \\
 & Graphiti & 24/24 & 0.440 & 0 \\
 & FIFO & 24/24 & 0.283 & 9 \\
 & No-recovery floor & 24/24 & 0.342 & 0 \\
 & Scratch & 24/24 & 0.062 & 6 \\
 & Belady & 24/24 & 0.060 & 6 \\
 & LRU & 24/24 & 0.053 & 12 \\
 & Scratch--LRU & 24/24 & 0.012 & 6 \\
 & Letta & 24/24 & 0.000 & 2 \\
\bottomrule
\end{tabular}
\end{table}

%% file: tables/tab_rq4_extended_architectures.tex
\begin{table*}[t]
\centering\small
\caption{\textbf{Expanded external-memory workload coverage.} Agent-executed end-to-end accuracy pooled over four models, three seeds, and the two registered capacities. All rows contain 24 effective outcomes. Running Maximum is a low-demand control, not a graded record-pressure workload.}
\label{tab:rq4-extended-architectures}
\begin{tabular}{llrrr}
\toprule
Workload & Architecture & Scored & Accuracy & Max-turn \\
\midrule
Stepwise Sum & A-MEM & 24/24 & 0.436 & 0 \\
 & Graphiti & 24/24 & 0.767 & 0 \\
 & Letta & 24/24 & 0.370 & 0 \\
 & Mem0 & 24/24 & 0.367 & 0 \\
\midrule
Stepwise Maximum & A-MEM & 24/24 & 0.942 & 0 \\
 & Graphiti & 24/24 & 0.934 & 0 \\
 & Letta & 24/24 & 0.008 & 0 \\
 & Mem0 & 24/24 & 0.861 & 0 \\
\midrule
Stepwise Top-$k$ & A-MEM & 24/24 & 0.429 & 0 \\
 & Graphiti & 24/24 & 0.440 & 0 \\
 & Letta & 24/24 & 0.000 & 2 \\
 & Mem0 & 24/24 & 0.289 & 4 \\
\midrule
Full Lookup & A-MEM & 24/24 & 0.949 & 0 \\
 & Graphiti & 24/24 & 0.986 & 0 \\
 & Letta & 24/24 & 0.177 & 0 \\
 & Mem0 & 24/24 & 0.512 & 0 \\
\midrule
Store--Recall & A-MEM & 24/24 & 0.959 & 0 \\
 & Graphiti & 24/24 & 1.000 & 0 \\
 & Letta & 24/24 & 0.370 & 0 \\
 & Mem0 & 24/24 & 0.996 & 0 \\
\midrule
Running Maximum (control) & A-MEM & 24/24 & 0.955 & 0 \\
 & Graphiti & 24/24 & 0.969 & 0 \\
 & Letta & 24/24 & 0.016 & 0 \\
 & Mem0 & 24/24 & 0.923 & 0 \\
\bottomrule
\end{tabular}
\end{table*}

%% file: tables/tab_rq2.tex
\begin{table}[t]
\centering\small
\caption{\textbf{Less retained state lowers accuracy, especially for
Stepwise Sum.} End-to-end accuracy (higher is better) as the number of retained
results $h$ decreases, with $m{=}16$ and recovery prohibited. Values are mean
$\pm$ standard deviation over three task seeds and three models
($3{\times}3{=}9$ runs); shading marks accuracy below 0.75. The ratio $m/h$
is a comparable pressure indicator across workloads, not a shared EIR formula.}
\label{tab:rq2}
\resizebox{\linewidth}{!}{%
\begin{tabular}{rrcccc}
\toprule
$h$ retained & $m/h$ & Store--Recall & Stepwise Sum & Stepwise Maximum & Stepwise Top-$k$ \\
\midrule
16 & 1.00 & $1.000\pm0.000$ & $1.000\pm0.000$ & $1.000\pm0.000$ & $0.861\pm0.246$ \\
12 & 1.33 & $0.875\pm0.018$ & \cellcolor{red!12}$0.292\pm0.058$ & $0.836\pm0.138$ & $0.868\pm0.089$ \\
 8 & 2.00 & $0.773\pm0.038$ & \cellcolor{red!12}$0.161\pm0.029$ & $0.770\pm0.178$ & \cellcolor{red!12}$0.667\pm0.223$ \\
 4 & 4.00 & \cellcolor{red!12}$0.667\pm0.043$ & \cellcolor{red!12}$0.089\pm0.007$ & \cellcolor{red!12}$0.570\pm0.327$ & \cellcolor{red!12}$0.431\pm0.311$ \\
\bottomrule
\end{tabular}
}
\end{table}

%% file: tables/tab_full_lookup_pressure.tex
\begin{table}[t]
\centering
\small
\caption{\textbf{Full Lookup pressure sweep.} Query accuracy under fixed
retention with recovery prohibited, pooled over four models and three seeds
($m=16$; 96 valid trials).}
\label{tab:full-lookup-pressure}
\begin{tabular}{rcc}
\toprule
$h$ & $m/h$ (operational) & Query accuracy \\
\midrule
16 & 1.00 & 0.997 \\
14 & 1.14 & 0.865 \\
12 & 1.33 & 0.781 \\
10 & 1.60 & 0.675 \\
 8 & 2.00 & 0.498 \\
 6 & 2.67 & 0.382 \\
 4 & 4.00 & 0.194 \\
 2 & 8.00 & 0.092 \\
\bottomrule
\end{tabular}
\end{table}

%% file: tables/tab_rescue.tex
\begin{table}[H]
\centering
\small
\caption{\textbf{RQ1 causal rescue.} Accuracy at Store--Recall window-miss
steps under no injection, the true missing result, or a length-matched sham
($m{=}8$, $h{=}4$; four models $\times$ three seeds = 12 trajectories per
arm). Each entry pools affected RECALL steps within those trajectories. The repair effect is
$\Delta_{\rm repair}={\rm accuracy}_{\rm exact}-{\rm accuracy}_{\rm sham}$.}
\label{tab:rescue}
\begin{tabular}{lcccc}
\toprule
Model & No injection & Exact restoration & Matched sham & $\Delta_{\rm repair}$ \\
\midrule
\texttt{Gpt56Reasoning} & 0.137 & 1.000 & 0.000 & $+1.000$ \\
\texttt{ClaudeOpus46}   & 0.176 & 1.000 & 0.000 & $+1.000$ \\
\texttt{ClaudeHaiku45}  & 0.000 & 1.000 & 0.000 & $+1.000$ \\
\texttt{Gpt41Mini}      & 0.140 & 1.000 & 0.000 & $+1.000$ \\
\midrule
Mean & 0.113 & 1.000 & 0.000 & $+1.000$ \\
\bottomrule
\end{tabular}
\end{table}

%% file: tables/tab_rq1_stepwise_max_rescue.tex
\begin{table}[t]
\centering\small
\caption{\textbf{Exact state restoration rescues Stepwise Maximum.} Full-trajectory accuracy under no injection, length-matched sham injection, and exact restoration at $m{=}8,h{=}4$. Each model--arm cell contains three seeds.}
\label{tab:rq1-stepwise-max-rescue}
\begin{tabular}{lrrr}
\toprule
Model & Control & Sham & Exact \\
\midrule
Claude Haiku 4.5 & 0.641 & 0.714 & 1.000 \\
Claude Opus 4.6 & 0.641 & 0.714 & 1.000 \\
GPT-5.6 & 0.646 & 0.714 & 1.000 \\
\midrule
Pooled mean & 0.642 & 0.714 & 1.000 \\
\bottomrule
\end{tabular}
\end{table}

%% file: tables/tab_rq3_cone.tex
\begin{table}[t]
\centering
\small
\caption{\textbf{Shared ancestors make naive reconstruction explode.} Mean
node evaluations needed to reconstruct one evicted result over three seeds
(lower is better). Atomic retrieval always costs one read. Memoized recursion
visits each required ancestor once, while naive recursion revisits shared
ancestors; ``speedup'' is naive cost divided by memoized cost.}
\label{tab:rq3-cone}
\begin{tabular}{lrrrr r}
\toprule
Topology & Atomic & \shortstack{Memoized\\recompute} & \shortstack{Naive\\recompute} & \shortstack{Memoization\\speedup} & Depth \\
\midrule
Chain ($f_{\rm in}{=}1$)    & 1 & 13.0 & $13$                 & $1\times$            & 13.0 \\
Balanced ($f_{\rm in}{=}2$) & 1 & 38.5 & $4.4{\times}10^{4}$  & $1.1{\times}10^{3}$  & 19.5 \\
Bushy ($f_{\rm in}{=}4$)    & 1 & 45.3 & $1.1{\times}10^{10}$ & $2.5{\times}10^{8}$  & 30.3 \\
\bottomrule
\end{tabular}
\end{table}

%% file: tables/tab_rq3_amplification.tex
\begin{table}[t]
\centering
\small
\caption{\textbf{Successful agents traverse efficiently; failures stop early.}
Model-executed reconstruction probes on \texttt{sum} DAGs. The harness performs
arithmetic, so reads isolate traversal behavior. Read overhead is reads divided
by the oracle minimum (1.0 is optimal; lower is not possible for a completed
probe). Early stops count terminations before reconstructing the target.}
\label{tab:rq3-amp}
\begin{tabular}{lccccc}
\toprule
Model & \shortstack{Success\\rate $\uparrow$} & \shortstack{Read overhead\\(successes; $1.0$ optimal)} & \shortstack{Redundant\\rereads $\downarrow$} & \shortstack{Off-cone\\reads $\downarrow$} & \shortstack{Early\\stops $\downarrow$} \\
\midrule
\texttt{Gpt56Reasoning} & 100\% & 1.000 & 0 & 0 & 0 \\
\texttt{ClaudeOpus46}   & 100\% & 1.000 & 0 & 0 & 0 \\
\texttt{ClaudeHaiku45}  & 100\% & 1.000 & 0 & 0 & 0 \\
\texttt{Gpt41Mini}      & 50\%  & 1.000 & 0 & 1 & 2 \\
\bottomrule
\end{tabular}
\end{table}

%% file: tables/tab_rq2_recovery_workloads.tex
\begin{table}[t]
\centering\small
\caption{\textbf{Expanded model-executed recovery probes.} Successful probes
reconstruct the target and use exactly the oracle-minimum reads; premature
probes stop before the required cone is complete. Counts pool both registered
fan-in conditions and all evaluated models.}
\label{tab:rq2-recovery-workloads}
\begin{tabular}{lrrr}
\toprule
Workload & Successful & Premature & Total \\
\midrule
Stepwise Sum & 32 & 8 & 40 \\
Stepwise Maximum & 18 & 6 & 24 \\
Stepwise Top-$k$ & 19 & 5 & 24 \\
\midrule
Total & 69 & 19 & 88 \\
\bottomrule
\end{tabular}
\end{table}

%% file: tables/tab_rq3_stepwise_max_behaviors.tex
\begin{table*}[t]
\centering\small
\caption{\textbf{Stepwise Maximum agent-managed behavior.} Accuracy and just-in-time (JIT) retrieval rate pooled over $h\in\{2,4\}$. Every model--architecture row contains all six registered seed--capacity outcomes.}
\label{tab:rq3-stepwise-max-behaviors}
\begin{tabular}{llrrr}
\toprule
Model & Architecture & Scored & Accuracy & JIT rate \\
\midrule
GPT-5.6 & No-recovery floor & 6/6 & 0.755 & -- \\
 & Scratch & 6/6 & 0.315 & 1.000 \\
 & LRU & 6/6 & 0.224 & 1.000 \\
 & Scratch--LRU & 6/6 & 0.089 & 1.000 \\
\midrule
Claude Opus 4.6 & No-recovery floor & 6/6 & 0.804 & -- \\
 & Scratch & 6/6 & 0.274 & 0.963 \\
 & LRU & 6/6 & 0.232 & 0.979 \\
 & Scratch--LRU & 6/6 & 0.089 & 0.940 \\
\midrule
Claude Haiku 4.5 & No-recovery floor & 6/6 & 0.654 & -- \\
 & Scratch & 6/6 & 0.227 & 0.881 \\
 & LRU & 6/6 & 0.182 & 0.738 \\
 & Scratch--LRU & 6/6 & 0.083 & 0.822 \\
\midrule
GPT-4.1 Mini & No-recovery floor & 6/6 & 0.401 & -- \\
 & Scratch & 6/6 & 0.075 & 0.990 \\
 & LRU & 6/6 & 0.250 & 1.000 \\
 & Scratch--LRU & 6/6 & 0.028 & 0.986 \\
\bottomrule
\end{tabular}
\end{table*}

%% file: appendix/deployment_results.tex
\section{VESTIGE: Information-Demand Measurement for Real Tasks}
\label{sec:deployment}
\label{sec:results-deployment}

VESTIGE measures task-relevant information demand for real tasks using
completed agent trajectories as observational data. These traces do not expose
exact continuation classes, so its
output is an operational retrospective proxy, not certified EIR or a causal
intervention. For software tasks, the VESTIGE semantic action graph combines observed
action--state dependencies with an independent reference patch: the trace
determines which file state was read, written, and later used, while files
changed by the reference patch identify goal-connected actions. Transactional
and shell benchmarks lack this artifact and use separate realized-state
adapters.

\subsection{The VESTIGE semantic action graph}
\label{sec:deployment-estimator}

Each raw log is normalized into turns $\tau_1,\ldots,\tau_T$ with explicit
file reads, writes, and estimated token costs. The semantic action graph links file state
to actions that read it and links write actions to the state they produce. To
capture information carried through the recorded work, a write also depends
conservatively on files read since the preceding write. Backward reachability
from observed accesses to reference-patch files identifies the recorded slice
connected to the task's reference solution.

The resulting graph provides observable necessity and utility signals for the
recorded execution: whether an action or artifact lies on a path to
solution-relevant activity, how long artifact state remains live, and whether
later goal-connected actions consume it. Aggregating these measurements across
runs estimates a task's demand profile; retaining them per run supports analysis
of access and reuse efficiency.

For a cut immediately before turn $t$, let $G$ be this backward slice, let
$s_t(p)$ be the file state visible at the cut, and let $u(s)$ be the final
downstream action in the slice that consumes state $s$. VESTIGE's live set is
\[
  L_t^{\mathrm{V}}=
  \{\,s_t(p):s_t(p)\in G\ \land\ u(s_t(p))\ge t\,\}.
\]
Its per-cut and peak retrospective demand estimates are
\[
  \widehat{\EIR}_{\mathrm{V}}(t)
  =\sum_{s\in L_t^{\mathrm{V}}}c(s),
  \qquad
  \widehat{\EIR}_{\mathrm{V}}^{\max}
  =\max_t\widehat{\EIR}_{\mathrm{V}}(t),
\]
where $c(s)$ is the estimated token cost of file state. The graph identifies
state used in the recorded continuation and connected to reference-solution
activity; it does not prove that the state is necessary for every correct
solution. A run with no observed reference-file access receives zero measured
demand, which indicates missing graph coverage rather than zero intrinsic
information demand.

\paragraph{Known-dependency validation.}
On 36 synthetic cases with known dependencies and an injected distractor,
VESTIGE retains all required files and excludes the distractor in every case.

\paragraph{Measurement scale and findings.}
We compute the VESTIGE signal for 72{,}562 software-agent trajectories:
4{,}988 SWE-bench Verified runs, 500 CoderForge runs, and 67{,}074 OpenHands
runs. Reference-file sink coverage is 92.0\%, 99.0\%, and 57.4\%, respectively.
Mean peak demand is 9.0k tokens on Verified, 73.1k on CoderForge, and 4.5k on
OpenHands. The corresponding failure AUCs are .541, .440, and .527, showing
that retrospective demand alone is not a strong failure classifier. In the
OpenHands clipping cohort, the demand--failure association is not significant
after adjustment for task difficulty (OR 1.016 per demand doubling, 95\% CI
$[0.992,1.041]$, $p=.191$).

\input{tables/tab_swe_retrospective}

\begin{table}[t]
\centering
\caption{Domain anchors for VESTIGE demand estimates. SWE uses the semantic action graph;
benchmarks without a reference solution retain separate realized-state
adapters and are not pooled with SWE token demand.}
\label{tab:eir-anchors}
\begin{tabular}{@{}lll@{}}
\toprule
Benchmark & State entity & Goal or anchor \\
\midrule
SWE-bench / SWE-rebench & versioned file & independent reference-patch file \\
$\tau$-Bench & typed database entity & reference task state \\
Terminal-Bench~2.0 & versioned shell path & realized trace only \\
\bottomrule
\end{tabular}
\end{table}

\input{appendix/deployment_extension}

\input{appendix/mechanistic_results}

%% file: tables/tab_swe_retrospective.tex
\begin{table}[t]
\centering
\caption{\textbf{VESTIGE semantic-action-graph results on completed software-agent runs.}
The graph combines observed dependencies with reference-patch sinks. Goal
access is the fraction of runs with at least one such sink.}
\label{tab:swe-retrospective}
\small
\begin{tabular}{@{}lrrrr@{}}
\toprule
Cohort & Runs & Goal access & \shortstack{Mean peak\\demand} & Failure AUC \\
\midrule
Verified & 4{,}988 & 92.0\% & 9{,}001 & .541 \\
CoderForge & 500 & 99.0\% & 73{,}135 & .440 \\
OpenHands/Qwen3 & 67{,}074 & 57.4\% & 4{,}460 & .527 \\
\bottomrule
\end{tabular}
\end{table}

%% file: appendix/deployment_extension.tex
\subsection{Software-agent trajectories: Verified, CoderForge, and OpenHands}
\label{sec:deployment-swebench}

We apply the VESTIGE semantic action graph to 4{,}988 real agent trajectories from
SWE-bench Verified~\citep{jimenez2024swe} produced by nine frontier models
(\texttt{gemini-3-pro-preview}, \texttt{claude-opus-4.6}, \texttt{claude-opus-4.5},
\texttt{gemini-3-flash-preview}, \texttt{minimax-m2.5}, \texttt{glm-5},
\texttt{gpt-5.2}, \texttt{claude-haiku-4.5}, \texttt{gpt-5-mini};
resolve rates 56--100\%).
The source contains 5{,}000 model--task configurations; 4{,}988 normalize into
eligible trajectories, while 12 are excluded because normalization does not
recover the required execution structure.
Unlike LACUNA, these trajectories expose no declared task graph and no closed-form
EIR. The operational estimate $\widehat{\EIR}_{\mathrm{V}}(t)$ combines
observed action--state dependencies with files edited by the reference patch.
The patch anchors one certified solution and is not assumed necessary for every
possible correct continuation.

\paragraph{Model contrast and pressure regime.}
Across nine models, mean context occupancy ranges from 1\% (\texttt{gpt-5-mini},
short trajectories, 56\% resolve) to 16\% (\texttt{gemini-3-flash-preview},
long exhaustive searches, 76\% resolve). These naive occupancy estimates are
low, but they do not establish $P<1$: neither exact real-task EIR nor effective
accessible capacity is certified, and supply can be limited when an observation
is first returned even if the accumulated history fits the model window. The
wide capability spread (56\%--100\% resolve)
captures substantial variation in failure behavior: models fail for
qualitatively different reasons---some quit too early, some search too broadly---
providing a stringent test of whether $\widehat{\EIR}_{\mathrm{V}}$ captures task difficulty
independently of model behavior.

\paragraph{Observed scaffold-level supply limits.}
We audit raw exports for scaffold-specific clipping markers rather than infer
availability from occupancy. In the publicly available OpenHands trajectory
logs, a truncated tool response contains the literal marker
\texttt{<response clipped>}. We treat this marker as evidence that the complete
tool output is unavailable in the recorded interaction, and therefore as a
proxy for tool-derived information supplied to the agent at that step. It is
not a reconstruction of the agent's full context: the exports omit per-call
prompt snapshots and do not reveal any hidden scaffold state or later
conversation condensation. All 5{,}000 mini-SWE-agent configurations cap each
command observation at 10{,}000 characters, retaining the first and last
5{,}000. This path fires 598 times in 265 trajectories; 135 events in 116
trajectories target a gold-patch file. All such events occur in
\texttt{gemini-3-pro}, which resolves all 500 tasks, so this cohort cannot
identify an effect on failure. In CoderForge, 244 file-view responses are
explicitly clipped across 203/500 trajectories; 116 events in 115 trajectories
target a gold-patch file. Per-call prompt size is monotone in all 500 runs,
providing evidence against silent history condensation in this scaffold, though
eight runs record context-window overflow errors. In OpenHands, 24{,}270 views
are clipped across 20{,}839/67{,}074 trajectories; 12{,}452 events in 12{,}114
trajectories concern files the agent later modifies. Modified-file clipping is
a behavioral anchor, not proof that omitted lines were normatively required.
No OpenHands-derived export contains explicit conversation-summary markers, so
absence of history condensation is not established. Resolve rates differ across clipping groups,
but task difficulty, exploration breadth, and model behavior confound those
observational contrasts. The audit establishes real output-level supply limits,
not that clipping caused failures.

We further restrict the gold-patch cohorts to source views whose requested
range demonstrably covers an old line deleted or replaced by the gold patch,
before the first detected write to that file. We then search later covering
views, also before that write, for the exact normalized old-line text. Pure
additions are excluded because their new lines did not exist to be viewed. In
Verified, 87 changed old-line instances across 30 runs are re-exposed after a
clipped view, while 29 instances across nine runs are not re-exposed before the
write or run end. All occur in \texttt{gemini-3-pro}, which resolves every
task. CoderForge contains 91 re-exposed instances across 30 runs and 34
non-re-exposed instances across six runs. It has only one run per task, so no
same-task outcome contrast is available. This refinement verifies omission of
specific gold-changed old-line text from retained tool output; it still does
not establish that the line was necessary for the agent's decision or that its
omission caused failure.

SWE-rebench provides an independent gold patch for all 6{,}306 OpenHands tasks
and all 67{,}074 runs. Applying the same mechanical test finds 15{,}404
re-exposed gold-changed old-line instances across 2{,}916 runs and 7{,}493
instances not re-exposed before the first detected write across 1{,}239 runs.
Restricting to exclusive trajectory groups without ambiguous line evidence,
raw task success is 48.5\% among 2{,}405 recovered runs and 35.8\% among 1{,}187
never-recovered runs, a $-12.7$ percentage-point difference (95\% CI
$[-16.1,-9.3]$). We retain all 3{,}592 runs in a logistic mixed model with a
random intercept for each of 676 tasks and fixed effects for total base-source
lines across gold-fix files, number of gold-fix files, and gold-patch additions
plus deletions. Ten tasks with unavailable base-file sizes use median
imputation with a missingness indicator. Model is constant because all
OpenHands runs use \texttt{Qwen3-Coder-480B}. The random-intercept association
is null (OR 0.89, 95\% credible interval $[0.67,1.19]$). A
correlated-random-effects sensitivity that additionally controls each task's
proportion of never-recovered runs is also null (OR 0.94, 95\% credible
interval $[0.71,1.26]$). To avoid relying only on random-effects assumptions,
a separate model estimates task success from runs outside both clipping groups
and again finds no association (OR 1.27, 95\% CI $[0.82,1.98]$, $p=.29$).
Leave-one-model-out estimation is unavailable because the cohort contains one
model. Recovery is itself post-clipping agent behavior, so even the adjusted
estimates are descriptive. The raw gap does not persist after adjustment for
task mix.

Figure~\ref{fig:trajectory-clipping-mechanism}A gives one trajectory-level
example using the literal marker in the public log rather than a simulated
context policy.

\paragraph{Demand and downstream behavior.}
Within the same 3{,}592 OpenHands clipping-classifier runs, we test whether
VESTIGE peak demand is associated with failure. Dividing by the common nominal
128k window only shifts the log scale, so coefficients are per demand doubling.
The association is small after controlling trajectory tokens (OR 1.022, 95\%
CI $[1.002,1.041]$, $p=.027$) and null after adding task difficulty estimated
from disjoint runs (OR 1.016, $[0.992,1.041]$, $p=.191$). VESTIGE demand uses the
completed trajectory and is not a clean pre-run causal exposure.

For mediation, the primary pre-run exposure is total source lines in the
gold-patch files, with median imputation and a missingness indicator for 56
runs. It predicts failure after trajectory-length adjustment (OR 1.30 per
doubling, $[1.10,1.55]$, $p=.003$), but not after independent task-difficulty
adjustment (OR 1.03, $[0.82,1.28]$, $p=.826$). Demand weakly predicts failure
to recover an omitted gold line (OR 1.19, $[0.99,1.42]$, $p=.059$), while
non-recovery does not independently predict failure in the joint model
(OR 1.41, $[0.92,2.17]$, $p=.115$). Task-cluster bootstrap standardization
gives an indirect risk difference of 0.19 percentage points from the 25th to
75th demand percentile (95\% interval $[-0.05,0.57]$; 200 resamples). An
exploratory rework proxy combining repeated commands, edit-application
failures, repeated edits, and submission after failed tests also shows no
mediated path (rework-to-failure OR 1.00, $[0.79,1.26]$, $p=.984$).
Thus neither adjusted VESTIGE demand nor the available downstream-behavior
measures support the proposed failure or mediation mechanism in this selected
clipping cohort.

\paragraph{Exploratory access behavior.}
We measure token distance from each read to the next read, overwrite, or trajectory
end. The inferential model replaces separate bins with
$x=\log_2(\text{distance}/500)$, using bin midpoints over the stable 0--32k
range. A task-fixed-effects Poisson model uses reread count as the outcome and
log token exposure as an offset; errors are clustered by run. It includes run
length, model indicators, failure, gold-needed status, and all lower-order
interactions. The sparse 32--64k needed-file bin contains one reread and is
excluded. The model contains 31{,}779 cells from 4{,}987 runs and all 500 task
fixed effects; one run has no usable file-version exposure.

The all-file failed-by-log-distance term is null (RR 1.018, 95\% CI
0.992--1.046, $p{=}0.174$). The pre-specified
failed$\times$needed$\times\log_2$(distance) term is negative (RR 0.951 per
doubling, 0.911--0.992, $p{=}0.020$; Table~\ref{tab:reread-distance}). Thus,
relative to resolved runs and to their behavior on other files, failed runs'
reread rate for gold-needed files falls by an additional 4.9\% per distance
doubling. Needed files are generally reread more often, so this is a relative
attenuation rather than an absence of retrieval.

The survival formulation is primary because a file that is never revisited
still contributes waiting-time exposure until run end; it does not condition on
eventual return. We nevertheless run three gap sensitivities. An
end-censored maximum gives OR 1.44, but can be mechanically inflated when failed
runs continue after useful work ends and is not treated as confirmatory. A
closed-gap analysis removes that tail but conditions on eventual reread or edit;
after task/model effects and run-length control its needed-file OR is 0.86. In a
joint closed-gap placebo model, needed-file OR is 0.87 and other-file OR is 0.81,
showing that this selected subset is not solution-specific.

The fair-clock sensitivity avoids both extremes by retaining nonreturns but
stopping every clock at the run's final edit of any file. The metric covers
4{,}574 runs and 261 discordant tasks. Raw medians are 2{,}326 tokens
(resolved) and 3{,}738 (failed). Conditional logistic regression with task and
model effects and $\log_2(1+\text{run tokens})$ gives OR 1.18 per gap doubling;
a 1{,}000-sample task-stratified run bootstrap gives [0.985, 1.43] (all fits
successful). Omitting run length gives OR 1.10, so adjustment does not hide a
large mediated effect here. The fair-clock direction agrees with the survival
model but is not independently significant. Neither analysis distinguishes a
model that retained the file, a model that lost track of it, and an agent that
followed a different reasoning path. They characterize recorded access behavior,
not context-caused failure.

\paragraph{Conservative information-use anomaly audit.}
We additionally search the 72{,}562 SWE traces for four exact log signatures:
tool-reported edit-application failure, identical command and output with no
intervening input-file write, exact edit repetition after a failed-test
candidate, and submission after an explicit failing-test result. Rules exclude
test-runner and environment failures and were manually reviewed on stratified
samples. Only 25 edit-application failures remain (25/25 reviewed events are
genuine tool failures), but they do not identify why the edit was stale. Exact
command/output repetition is common (77{,}164 events; 75/75 reviewed are exact),
yet the sample includes deliberate repro reruns and patch inspection, so it is
not a misuse label. The pipeline finds 110 repeated-edit candidates, but the
saved event representation lacks enough intervening test evidence to validate
them strictly. Only three submit-after-failed-test events remain after excluding
infrastructure failures; two are confirmed and one has truncated output. These
log-only signatures therefore do not provide an adequately powered,
high-specificity context-failure test.

\input{tables/tab_reread_distance}

\input{appendix/deployment_cross_domain}

%% file: tables/tab_reread_distance.tex
\begin{table}[t]
\centering
\small
\caption{\textbf{Recorded return rates differ by outcome and file relevance.}
The primary rate ratio comes from a task-fixed-effects Poisson model with log
token exposure as offset, run-clustered errors, run-length and model controls.
The fair-clock sensitivity ends at each run's final edit of any file; its
interval is a task-stratified run bootstrap.}
\label{tab:reread-distance}
\begin{tabular}{@{}llll@{}}
\toprule
Test & Estimate & 95\% CI & $p$ \\
\midrule
Failed $\times\log_2$ distance & RR 1.02 & [0.99, 1.05] & 0.174 \\
Failed $\times$ needed $\times\log_2$ distance & RR 0.951 & [0.911, 0.992] & 0.020 \\
Fair-clock needed-file gap, per doubling & OR 1.18 & [0.985, 1.43] & --- \\
\bottomrule
\end{tabular}
\end{table}

%% file: appendix/deployment_cross_domain.tex
\subsection{Cross-domain adapters: transactional entities and shell files}
\label{sec:deployment-cross-domain}
\label{sec:deployment-taubench}
\label{sec:deployment-tb2}

\paragraph{$\tau$-Bench entities.}
$\tau$-Bench~\citep{yao2024tau} contains persistent-state tool interactions but
no patch files. We normalize 318 frontier-model trajectories (226 retail and 92
airline) and treat typed database identifiers---users, orders, products, and
items---as state entities. Median peak later-used demand is four entities for
retail and two for airline; the 99th percentiles are eight and three. An
entity-counting task-side proxy has failure AUC 0.555 overall, but it is a
task-complexity association rather than exact EIR or evidence of context-caused
failure.

\paragraph{Terminal-Bench shell files.}
Terminal-Bench~2.0~\citep{merrill2026terminalbench} provides neither a normative patch
nor a declared dependency graph, but its trial directories retain nested
command/observation logs. Parsing shell commands into versioned file reads and
writes yields 166 analyzable trajectories from 178 trials; 12 have fewer than
three tool turns. Median peak later-used state is 22.5 files (11.5k estimated
tokens), the 90th percentile is 131 files (64.5k tokens), and the 99th
percentile is 455 files (226.5k tokens). Approximate path extraction and the
absence of a reference solution make this a realized shell-trace audit, not
exact EIR or normative demand.

Together these adapters show that hindsight demand can be reconstructed when a
trace exposes identifiable reads and writes. Their native units remain separate
from software-file token costs.

%% file: appendix/mechanistic_results.tex
\subsection{Mechanistic evidence across RQ1--RQ3}
\label{sec:mechanistic}

\paragraph{Evidence from execution logs.}
We inspect recorded replies and tool actions at the first missing-input
step and subsequent steps. In the Stepwise Sum example with $m=16$ and
$h=4$, the first missing input occurs at step 7, which needs $A_1$ and
$A_5$; only $A_5$ remains in context. These traces show how agents proceed
when an input is missing and how the resulting mistakes spread.

\paragraph{Continuing without acknowledging the missing input.}
A representative \texttt{Gpt56Reasoning} reply is simply
\texttt{ANSWER 7 3090}, although the correct answer is \texttt{6392}.
The recorded reply contains no recovery action or acknowledgment that
$A_1$ is missing. This establishes continued execution without visible
recovery, rather than what the model internally recognized.

\paragraph{Recognizing the loss but continuing anyway.}
The Claude traces make the missing input explicit. A representative
\texttt{ClaudeHaiku45} reply lists retained answers including
$A_4=5258$ and $A_5=1243$, then submits \texttt{3090}.
\texttt{ClaudeOpus46} states that $A_1$ is not resident before continuing.
In these examples, acknowledging the missing input does not lead to a
correct recovery; the agent proceeds using an invalid substitute.

\paragraph{Propagation through earlier answers.}
Once an incorrect result is retained, later steps may compute correctly
from incorrect inputs. We identify this pattern by applying the declared
operation to the agent's own retained answers: if it reproduces the
submitted answer using an incorrect earlier result, the step is classified
as a cascade error. This accounts for errors at steps with no new missing
input and supports the propagation analysis in
Section~\ref{sec:results-pressure}. The Stepwise Sum--Store--Recall
comparison in Figure~\ref{fig:rq2-cascade-gap} is consistent with this
mechanism, although the cross-workload difference is not itself a causal
estimate.

\paragraph{Stopping before recovery completes.}
In the recovery probes, \texttt{Gpt41Mini} can follow the correct
dependencies but stop before reaching the retained results needed to
reconstruct the target. Figure~\ref{fig:rq3-trace} contrasts completed
traversal with early termination in the original Stepwise Sum subset.
This differs from substitution: the agent begins a valid recovery but
does not finish it.

The traces distinguish three observable behaviors: continuing without
acknowledging the loss, acknowledging it without recovering the input,
and stopping during recovery. They explain how missing information can
lead to persistent errors or incomplete execution, complementing the
controlled restoration tests.